%% file: custom.tex
\pdfoutput=1

\documentclass[11pt]{article}
\usepackage[table,dvipsnames]{xcolor}
\usepackage[final]{acl}

\usepackage{inconsolata}
\input{math_commands.tex}

\usepackage{url}

\usepackage{colortbl}
\usepackage{adjustbox}
\usepackage{graphicx}
\usepackage{tikz}
\usepackage{caption}
\definecolor{darkgreen}{rgb}{0,0.5,0}
\definecolor{azureblue}{rgb}{0,0.5,1}
\definecolor{darkgreen}{rgb}{1,0,0}
\definecolor{color1}{HTML}{006EB8}
\definecolor{color2}{HTML}{009B55}
\definecolor{color3}{HTML}{00A99A}
\definecolor{color4}{HTML}{3C8031}
\definecolor{color5}{HTML}{006795}
\definecolor{color6}{HTML}{00AEB3}
\definecolor{mygray}{gray}{0.93}
\definecolor{mygreen}{HTML}{3FBC9D}
\definecolor{arsenic}{rgb}{0.23, 0.27, 0.29}
\usepackage[utf8]{inputenc}
\usepackage[T1]{fontenc}

\usepackage{url} 
\usepackage{booktabs}
\usepackage{amsfonts}
\usepackage{nicefrac}
\usepackage{microtype}
\usepackage{wrapfig}
\usepackage[nointegrals]{wasysym}
\usepackage{lipsum}  
\usepackage{times}
\usepackage{tikz}
\usepackage{latexsym}
\usepackage{microtype}
\usepackage{graphicx}
\usepackage{placeins}
\usepackage{subcaption}
\usepackage{bbm}
\usepackage{multirow}
\usepackage{enumitem}
\usepackage{tikz}
\usepackage{amsmath}
\usepackage{amssymb}
\usepackage{mathtools}
\usepackage{amsthm}
\usepackage{nccmath}
\usepackage{algorithm}
\usepackage{caption}

\usepackage{listings}
\usepackage[algo2e, ruled]{algorithm2e}
\SetKwComment{Comment}{$\triangleright$\ }{}
\SetKwProg{Init}{Initialize}{}{}
\definecolor{color_green}{HTML}{009B55}
\usepackage{multirow}
\usepackage{pifont}
\usepackage{graphicx} 
\usepackage{xspace}
\newcommand{\ourapproach}{\texttt{MMER}\xspace}
\newcommand{\pub}[1]{{\color{gray}{\tiny{[{#1}]\!}}}}
\newcommand{\foo}[2]{$\text{#1}\ _{(\text{#2})}$}

\definecolor{mygray}{gray}{.9}
\definecolor{ggray}{RGB}{127,127,127}
\definecolor{reda}{RGB}{192,0,0}
\definecolor{redb}{RGB}{217,148,143}
\definecolor{myyellow}{RGB}{190,144,0}
\definecolor{mygreen}{RGB}{80,100,40}
\definecolor{myblue}{RGB}{30,90,100}
\definecolor{tabhighlight}{HTML}{e5e5e5}

\usepackage{fontawesome5} 
\usepackage{seqsplit}

\title{Multi-Modality Expansion and Retention for LLMs through Parameter Merging and Decoupling}

\author{
    Junlin Li$^{1*}$,
     Guodong Du$^{1*}$,
     Jing Li$^1$\textsuperscript{\faEnvelope}, 
     \textbf{Sim Kuan Goh}$^2$,
     \textbf{Wenya Wang}$^3$, \\
     \textbf{Yequan Wang}$^4$,
     \textbf{Fangming Liu}$^5$,
     \textbf{Ho-Kin Tang}$^1$,
     \textbf{Saleh Alharbi}$^6$,
     \textbf{Daojing He}$^1$,
     \textbf{Min Zhang}$^1$ \\
    $^{1}$Harbin Institute of Technology, Shenzhen, China \quad $^{2}$Xiamen University Malaysia \\  
    $^{3}$Nanyang Technological University \quad
    $^{4}$Beijing Academy of Artificial Intelligence, China \\
    $^{5}$Peng Cheng Laboratory, China \quad
    $^{6}$Shaqra University, Saudi Arabia \\
    \texttt{leejunlin27@gmail.com} \quad   \texttt{jingli.phd@hotmail.com} \quad
}

\begin{document}
\maketitle

\begin{abstract}
        Fine-tuning Large Language Models (LLMs) with multimodal encoders on modality-specific data expands the modalities that LLMs can handle, leading to the formation of Multimodal LLMs (MLLMs).
        However, this paradigm heavily relies on resource-intensive and inflexible fine-tuning from scratch with new multimodal data.
        In this paper, we propose \textit{\ourapproach (Multi-modality Expansion and Retention)}, a \textit{training-free} approach that integrates existing MLLMs for effective multimodal expansion while retaining their original performance. 
        Specifically, \ourapproach reuses MLLMs' multimodal encoders while merging their LLM parameters.
        By comparing original and merged LLM parameters, \ourapproach generates binary masks to approximately separate LLM parameters for each modality.
        These decoupled parameters can independently process modality-specific inputs, reducing parameter conflicts and preserving original MLLMs' fidelity.
        \ourapproach can also mitigate catastrophic forgetting by applying a similar process to MLLMs fine-tuned on new tasks.
        Extensive experiments show significant improvements over baselines, proving that \ourapproach effectively expands LLMs' multimodal capabilities while retaining 99\% of the original performance, and also markedly mitigates catastrophic forgetting. 
    \let\thefootnote\relax\footnotetext{\faEnvelope~Corresponding author. $^*$ Equal contribution.}	
	\end{abstract}
	
	\section{Introduction}
	
	Large Language Models (LLMs)~\citep{2,91,87} have recently become a cornerstone in artificial intelligence due to their exceptional performance. 
	Building on LLMs, researchers~\citep{4,5}  integrate encoders for other modalities and use extensive modality-text data for alignment. 
	These synthesis are then fine-tuned to develop Multimodal LLMs (MLLMs), which excel at processing multimodal inputs. 
	This paradigm has led to the successful creation of numerous MLLMs across various modalities~\citep{7,8}.

     \begin{figure}[t]
		\centering
		\includegraphics[width=1\columnwidth]{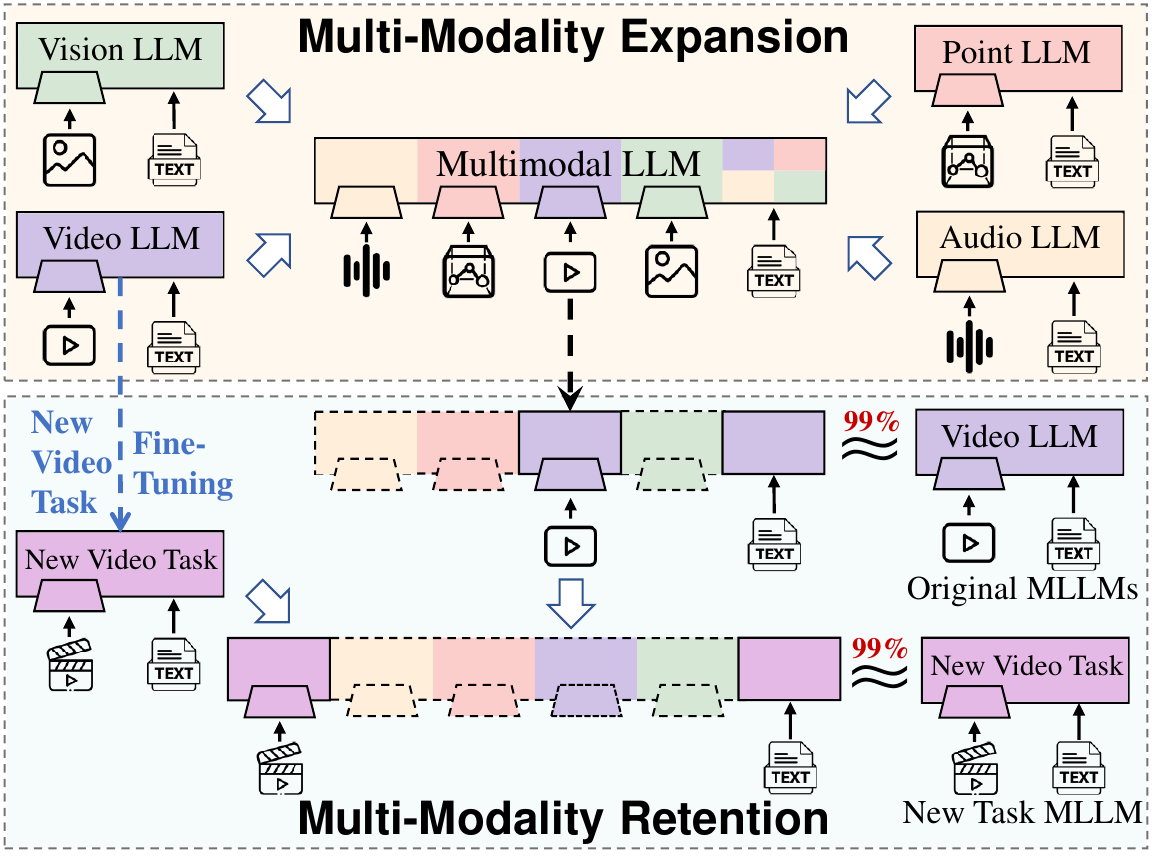}

		\caption{   The key ideas of \textbf{\ourapproach}. 
			Multi-Modality Expansion creates a versatile model from existing MLLMs via a \textbf{training-free, extensible} process. 
			Multi-Modality Retention reconstructs original or new task MLLMs to retain performance and mitigate catastrophic forgetting.
		}

		\label{fig:fig1}
	\end{figure}
    
	Most MLLMs specialize in dual modalities, including vision-oriented LLMs like LLaVA~\citep{5} and InternVL~\citep{9}, as well as video LLMs~\citep{10,11} and audio LLMs~\citep{12,13}. 
	Despite these advancements, there is a growing impetus to expand the modalities MLLMs can handle for diverse applications. 
	A straightforward method involves adding multiple new modality encoders~\citep{14,15} or employing a unified multimodal encoder~\citep{16}, followed by re-fine-tuning the MLLMs with fresh modality-text data.
	However, this method is resource-intensive and lacks flexibility, as it requires generating or acquiring high-quality multimodal instruction data~\citep{17,88} and fine-tuning from scratch.

	To overcome the aforementioned limitations, researchers have explored model merging for multimodal expansion in MLLMs~\citep{19,20}. 
	For instance, \citealt{18} proposed NaiveMC, a basic, training-free framework that 
    merges the LLMs of multiple MLLMs and combines their modality-specific encoders into the merged LLM.
    They further introduced the DAMC framework, which retrains MLLMs by separating modality parameters from language model parameters to mitigate parameter conflicts in the merged LLM.
	However, these two frameworks encounter a trade-off: NaiveMC is train-free but delivers lower performance, whereas DAMC requires training but yields better results. 

	In this paper, we propose a training-free approach named \ourapproach (Multi-Modality Expansion and Retention), which enables multimodal expansion while bypassing the above trade-off and retains the original performance (See Figure~\ref{fig:fig1}). 
	First, we merge the \textit{task vectors}~\citep{21}, which represent the difference between the fine-tuned and pre-train LLM parameters, into a merged task vector.
	Next, by comparing the \textit{Directional Congruence} and \textit{Dominant Significance} between the original and merged task vectors, we construct modality-specific binary masks. 
	These masks can approximately identify and decouple the original modality-specific parameters retained in the merged task vector.
	This strategy allows the merged MLLM to independently process non-textual modality data, using its decoupled parameters, thereby significantly reducing interference from other modalities.
	
	Furthermore, by re-adding a decoupled modality task vector into the base LLM parameters and integrating its corresponding encoder, we can reconstruct the near-original MLLMs. 
	This strategy can retain the original modalities' performance while saving storage space.
	Remarkably, since our \ourapproach approach is scalable, applying it to MLLMs fine-tuned on new tasks, along with multiple original MLLMs, yields a novel effect: effectively mitigating catastrophic forgetting. This approach enhances performance on new tasks without compromising previous ones by decoupling the new task's parameters from the original ones, thus preventing damage to the original parameters.

	We demonstrated the effectiveness of \ourapproach by composing four MLLMs (i.e., vision, audio, video, and point cloud) and conducted extensive experiments. 
    In multimodal tasks like MCUB~\citep{18}, \ourapproach significantly outperforms various baselines, confirming its ability to expand LLMs' multimodal capabilities without additional training.
	Moreover, we evaluated MLLMs reconstructed by \ourapproach on fourteen dual-modal tasks spanning four modalities paired with text. 
	The results reveal that they retain 99\% of their original performance.
	Lastly, \ourapproach proved resistant to catastrophic forgetting in single-task and cross-modal multi-tasks scenarios, effectively adapting to new tasks without undermining previous ones.

        Our work makes several \textbf{contributions}: 

        \begin{itemize}[noitemsep,nolistsep]
	\item We propose \ourapproach, a training-free approach for seamless multimodal expansion of LLMs through parameter merging and decoupling.
	\item We demonstrate two additional practical applications of the \ourapproach approach: retaining the performance of original MLLMs and mitigating catastrophic forgetting in MLLMs.
	\item We conduct extensive and rigorous experiments on various multimodal tasks across three scenarios, with confirm the effectiveness of the \ourapproach approach.
\end{itemize}

	\section{Related Work}
	\label{gen_inst}
	\paragraph{Multimodal Large Language Models.}
    Substantial researches~\citep{25,28,89} is dedicated to developing LLMs for multimodal inputs.
	Vision LLMs~\citep{26,4} excel in vision-language tasks by connecting visual encoders to LLMs, sparking a surge in dual-modality MLLMs. Other modalities, like audio and video, quickly followed suit~\citep{30,10}.
	Meanwhile, researchers explored unifying multiple modalities into a single LLM.
	ImageBind-llm~\citep{32} connects a multimodal encoder like ImageBind~\citep{33} to an LLM but relies solely on image-text data.
	OneLLM~\citep{16} improves this by aligning all modalities with language. 
	However, these methods cannot expand modalities due to the encoders have fixed input types.
	Other approaches connect multiple modality-specific encoders to an LLM, as seen in X-LLM~\citep{14}, MACAW-LLM~\citep{15}, which integrate encoders for vision, video, and audio. 
	However, these methods require high-quality multimodal data for joint training and still struggle with modality expansion.
	In contrast, \ourapproach provides an efficient, training-free solution for seamless multimodal expansion in LLMs.
\begin{figure*}[t]
		\centering
		\includegraphics[width=\linewidth]{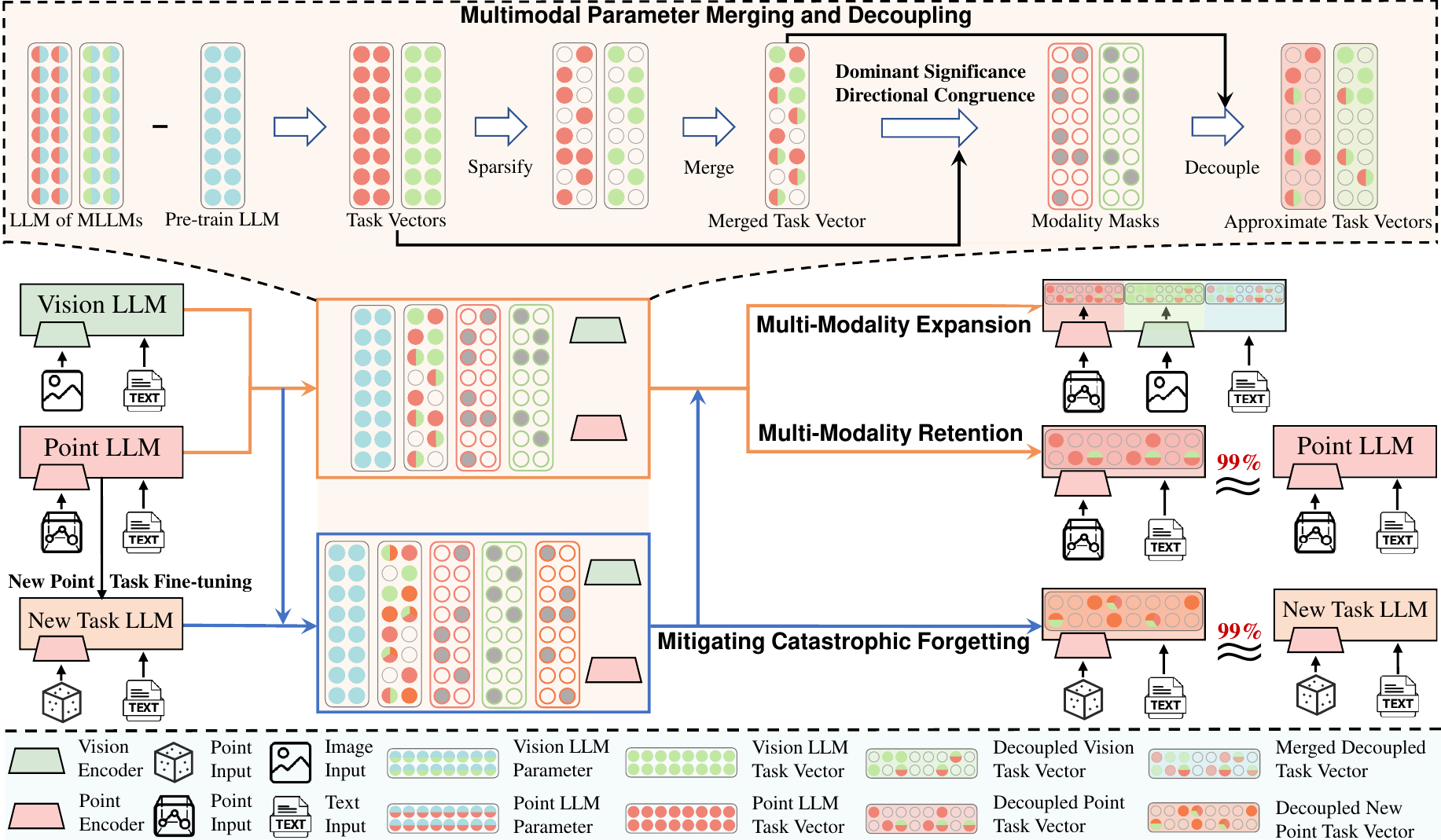}

		\caption{ The overview of \ourapproach, considering only the \textbf{Vision} and \textbf{Point Cloud} modalities \textbf{for clarity}.
  Each block corresponds to the same weight matrix, with empty blocks denoting zero value.
		``$\approx$'' signifies similar performance. 
		}
        \vspace{-10pt}
		\label{fig:fig2}
	\end{figure*}

	\paragraph{Model Merging and Model Composition.}
	Model merging~\citep{35,83} can improve single-task performance~\citep{37}, out-of-distribution generalization~\citep{39}, or combine the capabilities of multiple models~\citep{41}. 
    A basic method, TA~\citep{21} merges models by applying arithmetic operations on delta fine-tuned weights (i.e., task vectors), showing that such operations can yield comparable functional responses.
    Many subsequent methods have built upon this foundation.
	TIES~\citep{24} mitigates interference during merging by pruning redundant parameters and resolving sign conflicts, while DARE~\citep{43} achieves the same by randomly dropping and scaling parameters in a preprocessing step.
    Moreover,~\citealt{82} established the theoretical foundation for TA, showing that weight disentanglement is key to its success.
	Model merging further applies to multimodal models.
    ~\citealt{48} explored multimodal transformers merging for specific tasks. 
	Model Tailor~\citep{49} merges MLLMs to mitigate catastrophic forgetting.
	However, they do not explore the merging of MLLMs across modalities.
	To address this, the NaiveMC and DAMC frameworks~\citep{18} merge models to create a unified MLLM that inherits multiple modality capabilities, enabling seamless expansion. 
	However, one requires additional training, while the other delivers subpar performance.
        In contrast, \ourapproach enhances the multimodal expansion capabilities of MLLMs without extra training while retaining original performance and demonstrating resistance to catastrophic forgetting. Detailed comparison with related methods is in Appendix~\ref{app:NOVELTY}.

	\section{Methodology}
	\label{headings}
	
	In \ourapproach, we first merge the LLM parameters \(\{\theta_1, \theta_2, \dots, \theta_n\}\) from multiple MLLMs, all fine-tuned from the same LLM \(\theta_{\text{pre}}\), into a unified LLM. 
	However, such a merged model is prone to interference between modality-specific parameters, which can degrade the performance of representations.
        To handle this, we adopt a training-free parameter decoupling method that enhances the multimodal performance of the merged LLM while retaining the original performance. This is achieved by approximately decoupling modality-specific parameters within the merged parameter, ensuring independent processing of non-textual modality inputs.
    A visual workflow of \ourapproach is depicted in Figure~\ref{fig:fig2}.

	\subsection{Multimodal Parameter Merging and Decoupling}
	
    Since TA~\cite{21} showed the effectiveness of arithmetic operations on task vectors, which is further theoretically supported by~\citealt{82}, we apply these operations for parameter merging and decoupling.
	Specifically, we commence by employing the advanced model merging technique Ties~\citep{24} to merge \(\{\theta_1, \theta_2, \dots, \theta_n\}\).
	Ties first extracts the task vectors for each MLLM as \(\tau_{i,pre} = \theta_i - \theta_{\text{pre}}\), then refines them by selecting the Top\(K\%\) absolute values to filter out non-essential parameters. This results in sparse task vectors \(\tau_i\), which are then merged base on sign consistency to generate the merged task vector \(\tau_* = \textit{merge}( \sum_{i=1}^{n} \tau_i)\). Finally, the final merged LLM parameter is \(\theta_* = \theta_{\text{pre}} + \alpha \cdot \tau_*\), where \(\alpha > 0\) is a scaling factor calibrated by the validation set from target tasks. If these sets are unavailable, $\alpha$ is determined based on the model's general performance across tasks of each modality.

	Previous studies~\citep{50,44} show that most of the information from the task vectors is retained and embedded in the merged task vector \(\tau_*\).
	By comparing the original task vectors \(\tau_i\) with the merged task vector \(\tau_*\), we can identify relevant modality-specific parameter subsets from \(\tau_*\). 
	This enables the construction of modality-specific binary masks \(m_i\) to decouple and approximate each original task vectors \(\ m_i \circ  \tau_* \). These masks filter out irrelevant parameters and reconstruct the original model parameters \(\hat{\theta}_i\):
	\begin{equation}
		\hat{\theta}_i = \theta_{\text{pre}} + m_i \circ  \tau_* \approx \theta_i
	\end{equation}
	
	We construct the masks \(m_i\) by minimizing the Manhattan distance \(\ell_1^*\) between the reconstructed model \(\hat{\theta}_i\)  and the LLM \(\theta_i\) of original MLLMs:
	\begin{equation}
		\argmin_{m_i\in\{0,1 \}^P  }  \left| \hat{\theta_i} - \theta_i \right| = \argmin_{m_i \in \{0,1 \}^P } \left| m_i \circ \tau_* - \tau_i \right|  
        \nonumber
        \end{equation}
        \begin{equation}
        = \argmin_{m_i \in \{0,1 \}^P } \sum_{p=1}^P \left| m_i^{(p)} \circ \tau_*^{(p)} - \tau_i^{(p)} \right|
	\end{equation}
    
	where \( P \) represents the total number of parameters. 
    The rationale for using the Manhattan distance is analyzed in Appendix~\ref{manhattan}. 
	If the sign of \( \tau_i^{(p)} \) is inconsistent with that of \( \tau_*^{(p)} \), the masks \( m_i^{(p)} \) is set to 0 to avoid directional conflict.
	This step is referred to as \textbf{Directional Congruence}. 
	Conversely, when the sign of $\tau_i^{(p)}$ aligns with $\tau_*^{(p)}$ and $\left|\tau_i^{(p)}\right| 
	\geq \left| \tau_*^{(p)} - \tau_i^{(p)} \right|$, i.e., $\left|\tau_i^{(p)}\right| \geq 50\% \left|\tau_*^{(p)}\right|$, this indicates that $\tau_i^{(p)}$ is a dominant component of the merged parameter $\tau_*^{(p)}$. 
	Thus, $\tau_*^{(p)}$ can be approximated as $\tau_i^{(p)}$, so $m_i^{(p)}$ is set to 1, which we refer to as \textbf{Dominant Significance}.
	We further introduce a scaling factor $\lambda_i$ to refine this selection process, accommodating the varying numbers and modalities of original MLLMs, where a smaller $\lambda_i$ selects more parameters.
	The selection of $\lambda_i$ follows the same principle as $\alpha$, enabling the modality-specific inputs to be processed in parallel and independently.
	The final mask \(m_i\) is constructed by the following formula:
\begin{equation}
    m_i = \begin{cases}
        1 & \text{if} \ |\tau_i^{(p)}| \geq \lambda_i \cdot 50\% |\tau_*^{(p)}| \
  \text{and} \\
          & \quad  \text{sign}(\tau_i^{(p)}) = \text{sign}(\tau_*^{(p)}) \\
        0 & \text{otherwise}
    \end{cases}
\end{equation}
	

	\subsection{The \ourapproach Approach}
	We now comprehensively explain how the multimodal parameter merging and decoupling method enables multi-modality expansion, retention and addresses catastrophic forgetting in MLLMs.
	
	\subsubsection{Multi-Modality Expansion}
	Typical MLLMs consist of modality-specific components (i.e., multimodal encoders and alignment layers) and a base fine-tuned LLM. 
	Our \ourapproach approach disentangling these components, then applies the parameter merging and decoupling strategy to the fine-tuned LLMs of multiple MLLMs, producing a merged task vector \( \tau_* \), the pre-trained LLM parameter \( \theta_{\text{pre}} \), and \( n \) modality-specific binary masks \( m_i \). 
	The modality-specific components, including their weights, are reused directly, enabling the merged MLLM to seamlessly process all original modalities without losing functionality.

 \begin{figure}
\centering
\includegraphics[width=\linewidth]{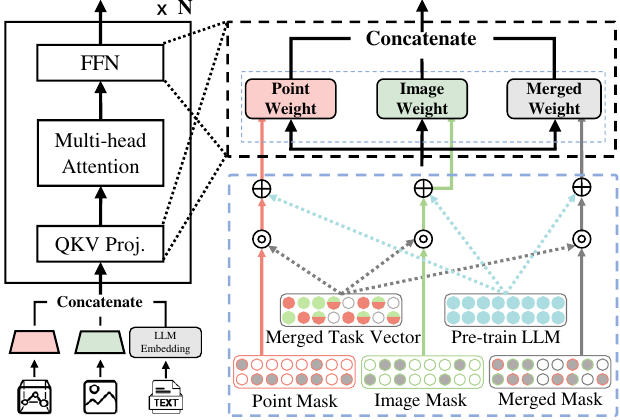}

\caption{ Details of \ourapproach's dynamic processing. \includegraphics[height=1em]{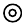} and \includegraphics[height=1em]{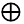} represent the Hadamard product and addition.}
\label{figure:attention}
\end{figure}
	As depicted in Figure~\ref{figure:attention}, upon receiving multimodal data, \ourapproach respectively encodes them into representation inputs \(X = [X_{M_1}, \dots, X_{M_n},  X_{t}]\), where $X_{M_i}$ and $X_{t}$ represent the modality-specific sequences and text sequences.
	\ourapproach then dynamically decouples the approximate modality-specific parameters \(\theta_{\text{pre}} +  m_i \circ \tau_* \). 
	This ensures that non-textual modality representations are processed independently with their respective parameters. 
	Text representations, on the other hand, are processed with the merged parameter \(\theta_{\text{pre}} +  \overline{m} \circ \tau_* \), where $\overline{m}$ is the average of all masks $m_i$. 
	For example, when representations progress to the attention mechanism at the \( l \)-th layer, \ourapproach decouples the modality-specific parameter from $W_{*,l}^Q$, the queries weights in the \( l \)-th layer from $\tau_*$, then: 
	
\begin{align}
    \mathbf{Q}_l = \left[ 
        {X}_{M_1,l} \left(m_{1,l}^Q \circ {W}_{*,l}^Q + {W}_{pre,l}^Q \right),  \right. \nonumber \\
     \left. \dots \ , \ {X}_{t,l} \left(\overline{m}_{l}^Q \circ {W}_{*,l}^Q + {W}_{pre,l}^Q \right) 
    \right]
\end{align}

	where ${W}_{pre,l}^Q$ denotes the queries weights in the \( l \)-th layer form $ \theta_{\text{pre}}$. 
	Afterward, \ourapproach sequentially decouples the modality-specific parameters for the keys and values in the \( l \)-th layer, and compute \( \mathbf{K}_l \) and \( \mathbf{V}_l \). 
	Finally, we carry out attention operation:
	\begin{equation}
		X^a_l = Attention(\mathbf{Q}_l, \mathbf{K}_l, \mathbf{V}_l)
	\end{equation}
	\begin{equation}
		[X^a_{M_1,l}, \dots, X^a_{M_n,l}, X^a_{t,l} ] = Split(X^a_l)
	\end{equation}
	
	Please note that the output representation should be partitioned by modality to match the input form. 
	Consequently, the final output of the attention mechanism at the  \( l \)-th layer is:
	
	\begin{equation}
		[{X}_{M_1,l}^o, \dots, {X}_{t,l}^o] = \left[ {X}_{M_1,l}^a  \left( m_{1,l}^O \circ {W}_{*,l}^O   + {W}_{pre,l}^O \right) \right. \nonumber 
        \end{equation}
        \begin{equation}
        \left. ,  \dots, {X}_{t,l}^a  \left(\overline{m}_{l}^O \circ {W}_{*,l}^O  + {W}_{pre,l}^O \right) \right]
	\end{equation}

	This procedure alleviates parameter conflicts across modalities, ensuring the merged MLLM retains fidelity when processing multimodal data.

	\subsubsection{Multi-modality Retention}
	Model merging and NaiveMC exhibit performance degradation (See Table~\ref{table2}) when handling modality-specific original tasks due to discrepancies between merged and original model parameters.
    However, \ourapproach circumvents this issue by approximately reconstructing the original MLLMs.
	This process involves decoupling the modality-specific task vector \(   m_i \circ {\tau_*}\), adding it to the pre-trained LLM \( \theta_{\text{pre}} \) to obtain the restored LLM \( \hat{\theta}_i = \theta_{\text{pre}} + m_i \circ {\tau}_* \), and then integrating the corresponding modality-specific components to reconstruct the final MLLM. 
	This strategy effectively mitigates parameter interference and retains original performance.

	\subsubsection{Mitigating Catastrophic Forgetting}
	
    Typically, fine-tuning MLLMs on new data improves performance on new tasks but often causes catastrophic forgetting on previous ones~\citep{23}.
    Drawing on the insight of Multi-modality Retention, \ourapproach can additionally mitigate catastrophic forgetting.
	We first fine-tune the corresponding original MLLM on the new tasks.
	Next, we apply the parameter merging and decoupling method to the fine-tuned MLLM, alongside all original MLLMs, generating a new merged task vector and binary masks.
	Finally, we reconstruct the corresponding MLLM in a targeted manner to handle different tasks.
	This enables \ourapproach to effectively adapt to new tasks without compromising previous ones, mitigating catastrophic forgetting.

	\section{Experiments Setup}
	\label{others}
	
	\subsection{Implementation}
    We explored \ourapproach across four MLLMs: Vision, Audio, Video, and Point Cloud LLMs.
	To ensure fairness and comparability, we fine-tuned these four MLLMs in the same environment, each based on Vicuna-7B-v1.5~\citep{51}, following previous works~\citep{18,20}. 
 	Details on experimental hyperparameters and fine-tuning can be found in Appendix~\ref{app:training}. 
    We evaluated performance based on evaluation scores or accuracy and performance retention, the latter as defined in Appendix~\ref{app:retention}.

    \input{tables/table1}
    \input{tables/table2}

    \subsection{Baseline Methods}
We compared \ourapproach with training-free methods: NaiveMC~\citep{18}, TA~\citep{21}, TIES~\citep{24}, and PCB-Merging~\citep{84}, where TA and TIES can substitute the merging strategy of NaiveMC for better performance. DARE~\citep{43} was integrated with these methods as it can complements them.
For multi-modality expansion experiments, we included training-based baselines: ImageBind-LLM~\citep{32} and X-InstructBLIP~\citep{20}.

    

    \subsection{Datasets and Tasks}
	
	In multi-modality expansion experiments, we evaluated multimodal tasks, including MCUB~\citep{18}, MUSIC-AVQA~\citep{54}, and ModelNet40~\citep{55} with images.
	For multi-modality retention experiments, we assessed fourteen dual-modal tasks spanning four modalities paired with text.
	Vision tasks include VQAv2~\citep{56}, GQA~\citep{57}, TextVQA~\citep{59}, VizWiz~\citep{60}, ScienceQA~\citep{61}, POPE~\citep{62}, and OK-VQA~\citep{63}. Audio tasks cover TUT~\citep{64}, VocalSound~\citep{65}, and Clotho~\citep{66}. Video tasks include MSRVTT~\citep{67} and MSVD~\citep{68}, and point tasks focus on ModelNet40~\citep{55} and Objaverse~\citep{58}.
	We evaluated \ourapproach's resilience to catastrophic forgetting on two new tasks, vision task Flickr30k~\citep{70} and audio task Clotho-AQA~\citep{69}.

	\section{Main results}

    \input{tables/table3}

	\textbf{Results on Multi-Modality Expansion.} As shown in Table~\ref{table1} , we observe the following:
	\textbf{(i)} Advanced training-free model merging methods improve the NaiveMC framework's performance, suggesting their effective application to the merging of MLLMs--a previously unexplored area.
	This also suggests considerable parameter conflicts in the merged MLLM, as these methods primarily focus on mitigating conflicts among merging parameters.
	\textbf{(ii)} Our \ourapproach approach significantly outperforms NaiveMC across all input combinations and tasks, demonstrating its effectiveness in extending multimodal capabilities and enhancing merged MLLMs' ability to manage modality combinations without additional training. 
	\textbf{(iii)} Furthermore, \ourapproach outperforms various baselines on nearly all tasks.
    This indicates that directly decoupling parameters after merging is more effective than merely reducing conflicts during the merging process.
    Lastly, the results for the original MLLMs are included in Appendix~\ref{ob_avqa}.

	\paragraph{\textbf{Results on Multi-Modality Retention.} } 
	The results in Table~\ref{table2}, reveal the following:
	\textbf{(i)} Interestingly, all methods show notable improvements on specific audio and point tasks. This likely due to these tasks are classification-based, whereas others involve captioning or QA tasks. 
	The original audio and point LLMs, not fine-tuned for classification tasks, fail to follow instructions leading to poorer performance.
	However, parameter merging may unlock their instruction-following ability, as the training data for other MLLMs included similar instructions. A detailed analysis is in Appendix~\ref{tisheng}.
    For fairer comparison, we also provide average performance trimming these tasks.
	\textbf{(ii)} Although NaiveMC enables multimodal expansion for handling multimodal tasks, its performance on original tasks substantially lags behind the original MLLMs.
	While varied model merging methods can somewhat alleviate this decline, the gap remains notable.
	In contrast, \ourapproach nearly retains the original performance. 
	For instance, \ourapproach achieves 99\% performance retention in the trimmed average.
	Detailed performance for each task is in Appendix~\ref{app:results}.
    
	\begin{figure}[t]
		\centering
		\includegraphics[width=\linewidth]{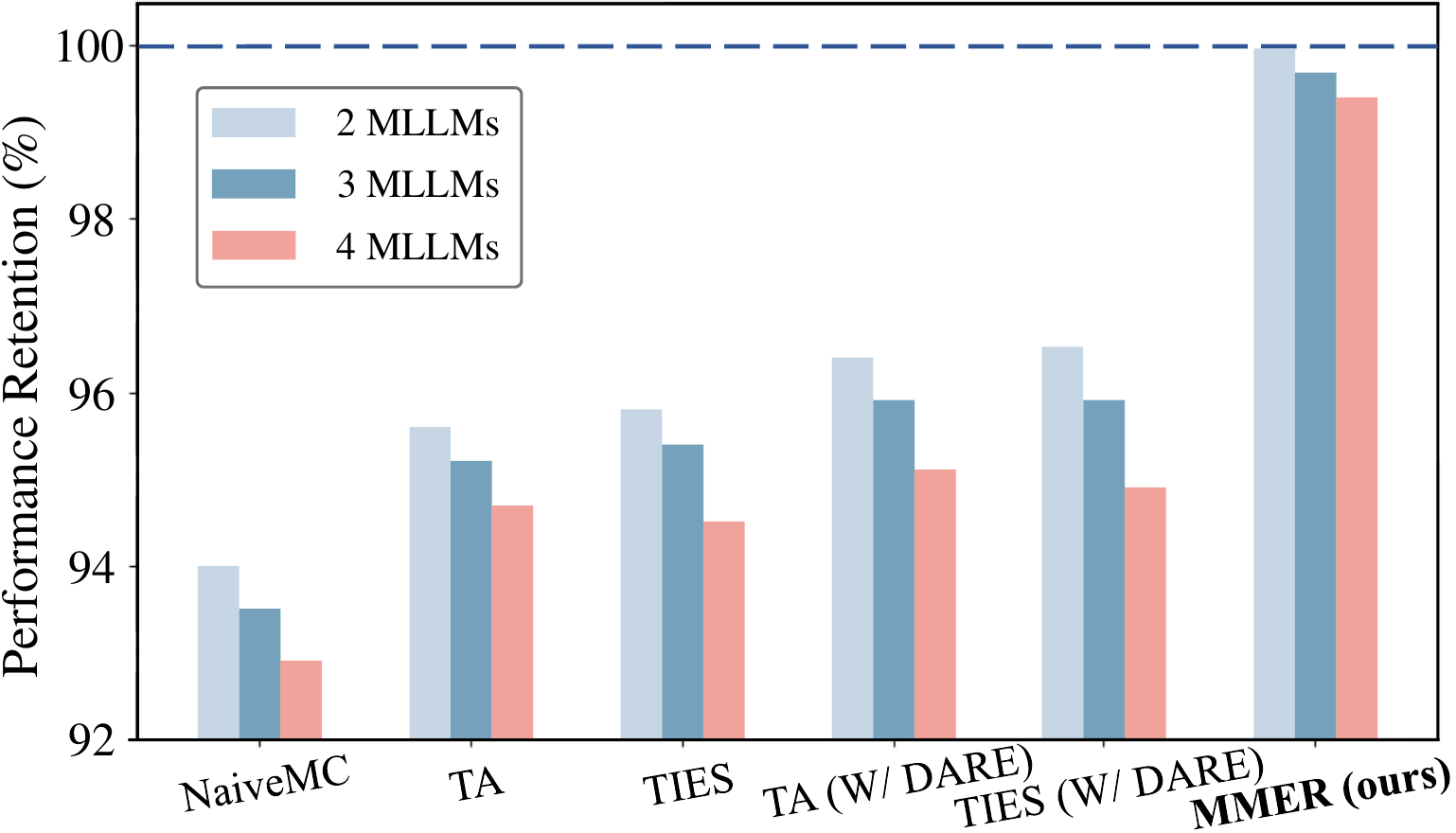}

		\caption{Performance retention vs. MLLMs quantity. 
		}

		\label{fig:fig4}
	\end{figure}

	\paragraph{Results on Mitigating Catastrophic Forgetting.}
    The results for both single-task and cross-modal multi-tasks scenarios are shown in Table~\ref{table3}.
	\textbf{(i)}
	Fine-tuning MLLMs boosts performance on new tasks but often compromises on previous ones.
     In contrast, \ourapproach, which additionally incorporates a fine-tuned MLLM (i.e., \ourapproach-Clotho-AQA or \ourapproach-Flickr30k), demonstrates strong robustness. 
     It maintains nearly original performance on previous tasks and adapts effectively to new ones, achieving results comparable to fine-tuned MLLMs.
	\textbf{(ii)} We further integrated both fine-tuned MLLMs into \ourapproach to showcase its performance in a cross-modal multi-tasks scenario. 
	As more MLLMs are integrated, \ourapproach continues to retain performance across new and previous tasks, though its ability to preserve performance slightly diminishes.
    Lastly, we compared \ourapproach with LoRA in Appendix~\ref{lora}. Detailed results for each task are provided in Appendix~\ref{app:results}.

\begin{figure}  
        \centering
    \includegraphics[width=0.83\linewidth]{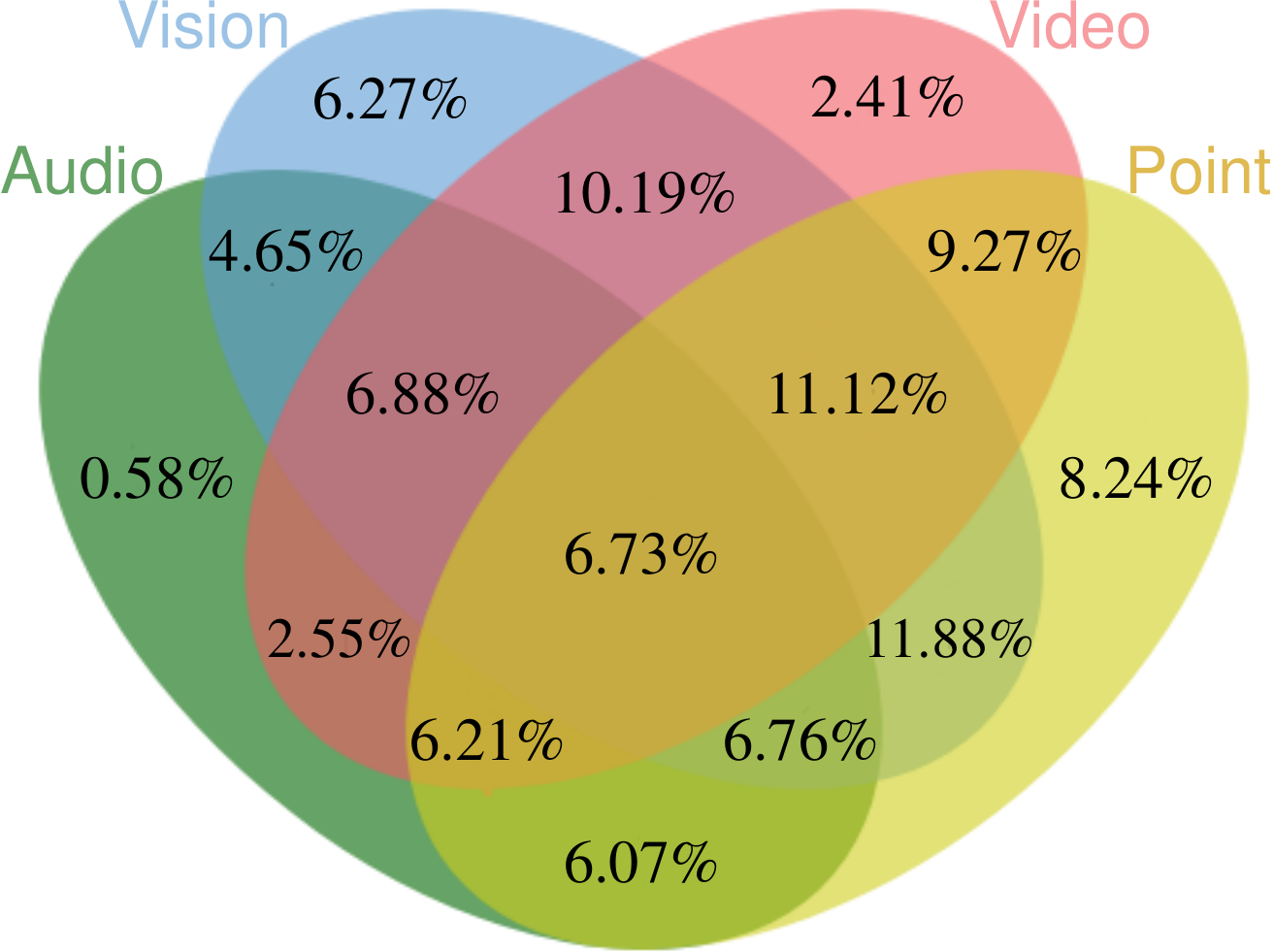}
    \caption{Parameters overlap across modalities.}

    \label{fig:venn}
\end{figure}

 \begin{figure*}[t]
		\centering
		\includegraphics[width=0.95\linewidth]{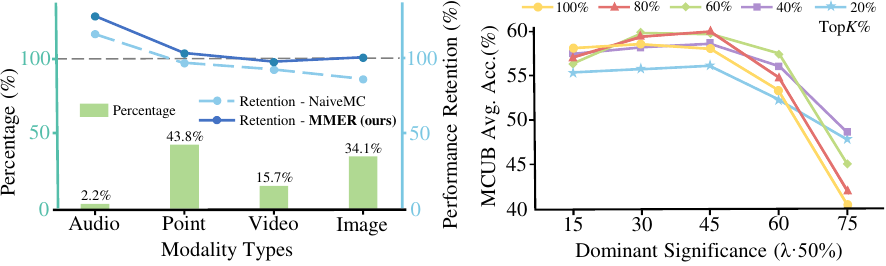}

	 \caption{\textbf{(Left).} The bar plots illustrate the percentage of parameters selected by masks, while the lines show performance retention of NaiveMC and \ourapproach across various dual-modal tasks.
		\textbf{(Right).} The lines depict the variations in MCUB average accuracy across different sparsity ratios (Top$K$\%) and Dominant Significance ($\lambda\cdot50\%$). 
	}
		\label{fig:fig3}

	\end{figure*}

	\section{Additional Results and Analysis}

	\paragraph{\textbf{Performance \& Storag vs. MLLM Quantity.}} Figure~\ref{fig:fig4} presents the performance retention of merging different numbers of MLLMs in retention experiments. 
    We can see that performance declines across all methods as more MLLMs are merged, indicating intensified parameter conflicts.
	Nevertheless, \ourapproach consistently outperforms other methods with only minor degradation, while other methods exhibit a noticeable drop when dealing with multiple MLLMs. 
	This highlights the robustness of parameter decoupling in mitigating conflicts.
    In terms of storage, \ourapproach significantly reduces costs compared to maintaining individual MLLMs while preserving similar performance and enabling multimodal expansion. Although it requires about twice the storage of model merging methods, it does not increase inference parameters and delivers notable performance improvements, striking an effective balance between the two approaches.
    Storage comparison details are in Appendix~\ref{app:storage}.
    
\paragraph{Parameters Overlap in Merged Task Vector.}

    Specifically, 40.43\%, 55.36\%, 64.49\%, and 66.28\% of audio, video, vision, and point parameters, are integrated into the merged task vector.
    The overlap between them shown in Figure~\ref{fig:venn}, reveals a severe conflict between parameters across modalities. 
    This underscores the need for \ourapproach to decouple key parameters and effectively mitigate conflicts.

	\paragraph{Modality-Specific Masks Analysis.} Figure~\ref{fig:fig3} (left) illustrates the percentage of parameters selected by different modality masks and compares the performance retention of \ourapproach with NaiveMC. 
	\ourapproach achieves performance close to or even exceeding the original levels, indicating that crucial modality-specific information is preserved after merging.
	Surprisingly, we find that the audio mask, retaining only 2.2\% of the parameters, still contributes to performance retention. 
	This phenomenon aligns with previous research~\citep{43}, which noted that ``\textit{Supervised fine-tuned language models tend to acquire excessively redundant delta parameters (i.e., task vectors).}'' Our results further confirm that this holds true for MLLMs as well.
    A detailed analysis and explanation are provided in Appendix~\ref{audio2.2}.

    \input{tables/table5}
	\paragraph{\textbf{Hyperparameters Analysis.}}Figure~\ref{fig:fig3} (right) examines the effects of the Top$K$\% hyperparameters and the scaling factor $\lambda$.
	Top$K$\% controls the sparsity of the original task vectors. 
	Excessive sparsity leads to marked performance degradation due to insufficient information in the sparse parameters.
	Conversely, insufficient sparsity fails to mitigate parameter conflicts, thereby hindering the decoupling of parameters.
	The effect of the scaling factor $\lambda$ is akin to Top$K$\%. 
    The scaling factor $\lambda$ regulates the extent of information the mask extracts from the merged task vector.
	If $\lambda$ is too high, the decoupled parameters lack effective information, leading to performance collapse. Conversely, if $\lambda$ is too low, irrelevant parameters persist, resulting in poor performance.
	In summary, Top$K$\% and $\lambda$ work in tandem to regulate the amount of effective information in the decoupled parameters. 

      \input{tables/table4}
	\paragraph{\textbf{Ablation Study.} }
	In Table~\ref{table5}, we begin with the original parameter decoupling strategy and systematically remove components to evaluate their effectiveness. 
	Removing Directional Congruence means selecting parameters based solely on Dominant Significance, i.e., $m_i = 1 \{\ |\tau_i| \geq 50\% \cdot \lambda_i |\tau_* | \}$. 
	Removing Dominant Significance retains parameters based only on the consistency of their signs, i.e., $m_i = 1 \{sign(\tau_i) = sign(\tau_*) \}$. 
	Table~\ref{table5} shows these components are crucial for optimizing performance. 
	Specifically, Directional Congruence is the most critical. 
	Without it, the decoupled parameters lose all original modality information and become nearly meaningless. 
	Next in importance is Dominant Significance. Without filtering out crucial parameters, irrelevant ones persist and disrupt the original parameters. 
	Finally, the scaling factor $\lambda$ also plays a role in further enhancing performance.

    \paragraph{\textbf{\ourapproach vs. Model Tailor.} }
    In Table~\ref{table4}, we compare our \ourapproach approach with the latest method for mitigating catastrophic forgetting in MLLMs within the same modality, since Model Tailor~\citep{49} is unable to accommodate new tasks across different modalities. 
The results show that \ourapproach consistently outperforms Model Tailor in both single-task and multi-tasks scenarios, highlighting its effectiveness. 
Furthermore, as the number of new tasks increases, \ourapproach maintains relatively stable performance, whereas Model Tailor exhibits a significant decline in performance on new tasks (i.e., from 91.69\% to 87.50\%), despite some improvement on previous tasks.
However, a minor drawback of \ourapproach is that its storage cost is approximately twice that of Model Tailor. 
Nonetheless, as the number of new tasks grows, \ourapproach's practicality becomes more pronounced, making it a more viable solution in scenarios where balancing performance and storage efficiency is crucial.

	\section{Conclusion}
	In this paper, we propose \ourapproach, a training-free method that resolves the dilemma of multimodal expansion for LLMs: costly retraining or suboptimal performance.
	\ourapproach retains the multimodal encoders of existing MLLMs, merges their LLM parameters, and constructs binary masks to decouple modality-specific parameters. 
	This mechanism enables independent handling of modality-specific inputs, reducing parameter conflicts.
	Besides, \ourapproach can reconstruct original MLLMs, effectively retaining their performance and mitigating catastrophic forgetting.
    We conducted extensive experiments and analyses to validate the effectiveness and robustness of our \ourapproach approach.

\section*{Acknowledgements}
This work was supported by National Science Foundation of China (62476070), Shenzhen Science and Technology Program (JCYJ20241202123503005,  \seqsplit{GXWD20231128103232001,~ ZDSYS20230626091203008,~ KQTD2024072910215406)}  and Department of Science and Technology of Guangdong (2024A1515011540).
This work was also supported in part by the Major Key Project of PCL under Grant PCL2024A06 and PCL2022A05, and in part by the Shenzhen Science and Technology Program under Grant RCJC20231211085918010.

\section*{Limitations}

We have focused exclusively on four commonly used modalities, leaving out a thorough analysis of the full range of potential modalities.
Additionally, finding multiple existing MLLMs with the same architecture across modalities is currently challenging, and due to limited computational resources, experiments on larger-scale MLLMs are constrained.
Finally, although our \ourapproach approach does not increase inference parameters, the storage cost is twice that of the base model.

\section*{Ethical Considerations}
Our research is conducted using publicly available and safe datasets and models. However, we explicitly acknowledge that the applicability of our \ourapproach approach and findings may be limited to datasets or domains similar to those studied. The performance of our approach on other specific datasets or domains remains uncertain, and there may be potential risks when applying it to privacy-sensitive or high-risk scenarios. Therefore, caution is advised, and thorough verification is necessary to ensure the method generates accurate and reliable results in such contexts.


\bibliography{custom}

\newpage

\appendix
	\section{Novelty and Contributions}
        \label{app:NOVELTY}
        Our research aims to achieve training-free multi-modality expansion and retention for LLMs through parameter merging and decoupling. 
        We conduct a comparative analysis with existing relevant methods to demonstrate the innovation of our \ourapproach approach.

        \paragraph{\textbf{Comparison with NaiveMC and DAMC frameworks.}}
        Our \ourapproach approach is based on the NaiveMC framework~\citep{18} and employs a parameter dynamic decoupling strategy similar to that of the DAMC framework~\citep{18} to mitigate parameter conflicts in the merged MLLM. 
        However, there are several key differences:
        \begin{enumerate}
            \item Compared to the NaiveMC framework, our \ourapproach approach effectively enhances the multimodal performance of the merged MLLM.

            \item Compared to the DAMC framework, our \ourapproach approach employs a training-free parameter decoupling strategy instead of separating parameters during the initialization training of the MLLMs and achieves similar results. 
            Additionally, \ourapproach is additional compatible with full-parameter fine-tuned MLLMs, whereas DAMC is restricted to parameter-efficient fine-tuned MLLMs.

            \item Compared to the NaiveMC and DAMC frameworks, our \ourapproach approach retains the performance of the original MLLMs while also providing additional capabilities to mitigate catastrophic forgetting.
        \end{enumerate}
        Our \ourapproach approach integrates the strengths of the NaiveMC and DAMC frameworks, while additionally providing original performance retention capabilities.
        
        \paragraph{\textbf{Comparison with training-free model merging methods.}}
       Training-free model merging methods, such as TA~\citep{21}, TIES~\citep{24}, PCB-Merging~\citep{84}, and DARE~\citep{43}, are primarily designed for merging models with identical architectures. 
       Consequently, they must be combined with the NaiveMC framework to achieve multi-modality expansion for LLMs. 
       These methods alleviate parameter conflicts in merged MLLMs to some extent, leading to performance enhancement.
       However, their overall effectiveness, both in terms of multimodal performance and retention of original performance, falls significantly short compared to our \ourapproach approach.

        \paragraph{\textbf{Comparison with alignment and fine-tuning methods.}}
        Compared to methods~\citep{14,15,16} that achieve multimodal expansion for LLMs by adding multiple new modality encoders or employing a unified multimodal encoder followed by alignment and fine-tuning, the advantages of our \ourapproach approach are clear.
        \ourapproach can effectively reuse a large number of MLLMs from the open-source community and merge them enabling multimodal expansion without the need for extensive resources and time spent on training models and constructing high-quality modality instruction data.

        \paragraph{\textbf{Comparison with TALL-masks.}}
        TALL-masks~\citep{44} is an information localization algorithm that, similar to our approach, compresses original parameters and subsequently approximates their restoration. However, there are several key differences:
        \begin{enumerate}
            \item From an algorithmic perspective, TALL-masks overlooks the Consistency of original and merged parameter signs. In contrast, we have addressed this aspect and demonstrated its effectiveness in our ablation experiments (See Table~\ref{table5}).

            \item In terms of application scenarios, our \ourapproach applies parameter merging and decoupling to the multimodal expansion for LLMs, enhancing their multimodal capabilities. 
            Additionally, we utilize \ourapproach to mitigate catastrophic forgetting. These aspects are not considered by TALL-masks.

            \item Regarding the models utilized, the models used in our \ourapproach approach are the 7B MLLMs across various modalities, while TALL-masks is applied to relatively smaller models within the same modality, such as T5~\citep{71} and ViT~\citep{72}.
        \end{enumerate}
        
                \input{tables/table_mllms}       
        \input{tables/table_hyper}  
        
	\section{Implementation and Experimental Details}

         All our experiments are conducted on an NVIDIA 8×A800-SXM4-80GB machine.
         \subsection{Performance Retention}
        \label{app:retention}
        Considering the varying modalities of each original MLLM and the different evaluation metrics for distinct tasks, we provide performance retention in our results to validate the method's capacity to retain original performance.  The definition is as follows:
         \begin{equation}
    \text{PR} =  \frac{1}{T} \sum_{t=1}^{T} \frac{\underset{x\sim\mu_t}{\text{metric}}\left[f_\text{method}(x)\right]}{\underset{x\sim\mu_t}{\text{metric}}\left[f_\text{original}(x)\right]}
\end{equation}
        where PR stands for performance retention and the ``metric'' refers to various evaluation metrics, such as accuracy and captioning scores(e.g., BLEU, ROUGE)~\citep{86,85,90}.

        \subsection{Implementation Details of Parameter Merging and Decoupling Process and Original Fine-tuned MLLMs}
         \label{app:training}
         For the parameter merging and decoupling process, we set Top\textit{$K$} to 80\%, while $\lambda$ was calibrated according to the modality. We did not set the value of $\alpha$ as we did not use the merged MLLM merging by TIES in \ourapproach.
         For fine-tuning the original MLLM, we used the same training data and components of each MLLM across the four modalities following NaiveMC~\citep{18}. More details are presented in Table~\ref{table:mllms}.
        We adopted similar hyperparameters following previous works~\citep{18, 52,20,10,53}. During the alignment stage, only the parameters in the connectors were trainable. In the fine-tuning stage, we tuned all connector parameters and base LLM parameters. For training efficiency, we utilized DeepSpeed Zero Optimization Stage 3. Detailed data are presented in the Table~\ref{table:hyper}.

        \subsection{Baseline  Details}
        In this section, we provide a detailed overview of the six baselines included in our experiments: 
        \begin{itemize}
            \item \textbf{Original MLLMs} means that each MLLM is evaluated on its corresponding modality tasks to demonstrate its original performance, but they cannot perform cross-modal tasks simultaneously.
            
            \item \textbf{NaiveMC framework}~\citep{18} combines modality-specific encoders from multiple MLLMs into the merged LLM, which is obtained by averaging the parameters of multiple LLMs from these MLLMs. The averaging merging strategy can be replaced by other model merging methods.
            
            \item \textbf{TA}~\citep{21} initially defines the concept of \textit{task vector} and employs arithmetic operations for model merging, model forgetting, and support multi-tasks learning, etc. 
            The final model is formed by scaling and adding task vectors to the initial model, represented mathematically as $\theta_m = \theta_\textrm{init} + \lambda \cdot \sum_{t=1}^n \tau_t$.
            
            \item \textbf{TIES}~\citep{24} improves upon TA~\citep{21} by further mitigating parameter interference.
            It first prunes redundant parameters to retain the most important ones. When encountering conflicts in parameter signs during merging, it selects and merges parameters with the majority sign while ignoring those with minority signs.

            \item \textbf{DARE}~\citep{43} proposes a preprocessing step to address parameters conflict. This method randomly discards the majority of the delta parameters while scaling the remaining ones by \( \theta' = \theta \cdot (1/(1-p)) \) where \( p \) is the proportion of dropped delta parameters. 

            \item \textbf{Model Tailor}~\citep{49} identifies the key parameters fine-tuned on the new tasks within the MLLM and integrates them into the original MLLM, thereby retaining the performance on previous tasks while adapting to new tasks.

        \end{itemize}

\section{Storage Cost Calculation}
  \label{app:storage}
\begin{figure}[t]
		\centering
		\includegraphics[width=\linewidth]{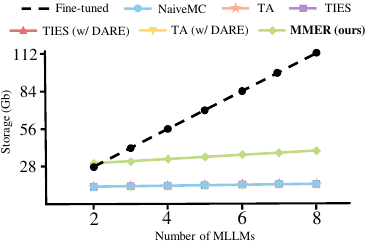}
  \vspace{-15pt}
		\caption{Storage cost vs. Number of MLLMs. 
		}
  \vspace{-9.3pt}
		\label{fig:storage}
	\end{figure}

  As shown in Figure~\ref{fig:storage}, although model merging methods maintain low storage costs that remain constant regardless of the number of merging MLLMs, their lower performance may constrain their practical applicability. 
	In contrast, maintaining individual MLLMs preserves strong performance for their respective modalities but fails to achieve multimodal expansion and results in linear growth in storage costs.
	Our \ourapproach approach strikes an effective balance between these approaches. It enables multimodal expansion while retaining nearly 100\% of the original MLLMs' modality capabilities and provides additional resilience against catastrophic forgetting.

  Additionally, we provide the calculation of storage costs for \ourapproach approach and the relevant methods mentioned above.
  Let $N$, $P$, $P'$, and $P^*$ represent the number of original MLLMs, the total parameters of the LLMs, the number of the modality-specific component parameters, and the number of additional trainable parameters of parameter-efficient fine-tuning methods, respectively.
    Assuming each float parameter occupies 32 bits, the storage cost for these methods across $N$ original MLLMs is calculated as follows:
    \begin{itemize}
    \item Original fine-tuned models: $32N(P+P')$. 
    $32(P+P')$ represents the number of parameters contained in a single MLLM.
    \item NaiveMC framework: $32P+32NP'$. Stores a merged LLM and $N$ modality-specific components.
    \item DAMC framework: $32P+32NP'+2N(32P^*)$. Stores a merged LLM and $N$ modality-specific components. $2N(32P^*)$ represents the need to store an additional $2N$ trainable parameters of parameter-efficient fine-tuning methods for parameter separation.
    \item NaiveMC wit TA / TIES / DARE: $32P+32NP'$. Same as the NaiveMC framework.
    \item \ourapproach: $64P+32NP'+NP$. $64P$ is for storing the parameters of a base LLM and a merged task vector, while $32NP'$ indicates $N$ modality-specific components. Additionally, $NP$ denotes the storage for
$N$ binary masks.
\end{itemize}
    
  \section{{More Analysis}}
    \subsection{Rationale for Using The Manhattan Distance}
    \label{manhattan}
    Firstly, we do not adopt methods like Fisher~\citep{80} or Regmean~\citep{81}, which require additional gradient-based computations to obtain the information matrix, as they demand substantial computational resources or data. Inspired by TIES~\citep{24} and DARE~\citep{43}, which propose that \textit{``Supervised fine-tuned language models tend to acquire excessively redundant delta parameter''}, we aim to decouple the most critical parameter of each modality from the merged task vector so that the decoupled parameters are as close as possible to the original task vectors.
    
    Based on the aforementioned concept, we decided to use a binary mask matrix to directly mask out irrelevant parameters in the merged task vector, retaining only the key information related to each modality.
    We chose to use the Manhattan distance to optimize the mask mainly due to its mathematical properties and its promotion of sparsity in high-dimensional parameter spaces. 

    \input{tables/l1l2}

    In particular, since most of the delta parameters are redundant, this implies that most elements in the mask should be zero, with only a few elements set to 1. By minimizing the Manhattan distance, we can easily achieve this goal because the gradient of parameter updates with respect to the Manhattan distance is constant. This makes it more likely to penalize smaller non-zero parameters and drive them to zero, thus encouraging the sparsity of the mask. Moreover, these smaller non-zero parameters are often redundant~\citep{24}, which are the ones we wish to mask out.

    Furthermore, Manhattan distance directly measures the element-wise difference between the merged task vector and the original task vectors. This comparison can precisely capture which parameters have undergone significant changes during fine-tuning and which parameters are irrelevant noise.    
    Finally, We conducted both multi-modality expansion and retention experiments by replacing the Manhattan distance with the Euclidean distance. The results presented in the Table~\ref{l1l2} validated the effectiveness of using Manhattan distance.

    \subsection{{Modality-Specific Masks Further Analysis}}
    \label{audio2.2}
     We construct the audio mask by comparing the merged task vector with the original audio MLLM task vector. Thus, the audio mask selecting only 2.2\% of the parameters reflects the significant difference between these two task vectors. Next, we analyze why the remaining 97.8\% of parameters were not selected. There are two possible reasons for the unselected parameters:
    \begin{enumerate}
        \item The signs of $ \tau_*^{(p)}$ and $ \tau_{audio}^{(p)}$ are opposite.
        \item The signs of $ \tau_*^{(p)}$ and $ \tau_{audio}^{(p)}$ are the same, but the magnitude of $ \tau_{audio}^{(p)}$ is too small.
    \end{enumerate}
    We examined the percentage of $ \tau_{i}^{(p)}$ whose signs align with those in the merged task vector and the average magnitude of $ \tau_{i}^{(p)}$ across four modalities, the results are shown in Table~\ref{table_tvs}.

    \input{tables/table_tvs}
    
    It is evident that the direction mismatch is not the primary cause, as the percentage differences in directional alignment across the four modalities are relatively small. However, we found that the magnitude of the audio task vector is significantly smaller than those of the other modalities. This indicates that the original audio MLLM is highly similar to the pre-trained LLM. As a result, the merged model (97.8\% of the parameters from the pre-trained LLM with 2.2\% of the parameters activated by the audio mask from the merged task vector) only needs to activate 2.2\% of the key parameters to retain its audio performance.

\input{tables/table_lora}

    \subsection{{Analysis of Performance Improvement in Multi-Modality Retention Experiment}}
    \label{tisheng}
    
    Firstly, the performance gain is not due to the removal of redundant parameters. In general, as more parameters are removed, performance tends to degrade~\citep{24,43}. This trend was also evident in our analysis (see Figure~\ref{fig:fig3} (right)), where increasing the Dominant Significance $\lambda$·50\% resulted in a reduction of selected parameters for each modality, leading to a gradual decline in performance.

    So, what accounts for the performance improvement? We hypothesize that the parameters selected by the mask overlap with parameters from other modalities. To explore this further, we analyzed the overlap of the parameters selected by the audio mask with those from other modalities. We found that 41.7\% of these parameters do not overlap with any other modality, while 23.2\%, 21.1\%, and 22.1\% overlap with the video, vision, and point modalities, respectively.

    It is possible that the model benefits from additional knowledge embedded in these overlapping parameters, such as prior knowledge or instruction-following capabilities. To validate this hypothesis, we replaced the overlapping parameters with the original audio task vector and conducted experiments on three audio tasks, yielding results of 24.71 (97.6\%) / 24.32 (98.4\%). Notably, the performance improvement was lost, which confirms the validity of our analysis.

\section{Detailed Results and Extended Experiments}

    \subsection{Mitigating Catastrophic Forgetting Experiments}

    \paragraph{\textbf{\ourapproach vs. LoRA.} }
    \label{lora}
    
    {We fine-tuned a LoRA adapter on original vision MLLM for Flickr30k, with the detailed results presented in Table~\ref{table_lora}.
    The results show that LoRA improves performance on target tasks but inevitably leads to a decline in performance on previous tasks, although this decline is less severe compared to full-parameter fine-tuning. In contrast, our \ourapproach approach outperforms LoRA on target tasks, while causing almost no degradation in previous tasks. However, this comes at the cost of increased storage overhead. Both approaches have distinct advantages and disadvantages, enabling users to select the most suitable method based on their specific requirements.}
    
    More importantly, our approach addresses an additional application scenario. In the open-source community, models are typically categorized into adapter-based models and full-parameter fine-tuned models. While the former can be easily integrated into existing models, the latter lacks such adaptability. Our approach bridges this gap by providing a solution to seamlessly incorporate full-parameter fine-tuned models.

    \input{tables/table_unimodal}
    \input{tables/table2_detail_I}
    \input{tables/table2_detail_P}
    \subsection{Further Results of Multi-Modality Expansion Experiments}
    \label{ob_avqa}
    {
    We supplemented the results of four original unimodal models on the multimodal tasks for a fairer comparison. 
    Since MCUB cannot be evaluated using unimodal models, we excluded it from the analysis.
    As shown in Table~\ref{table_unimodal}, we observe that \ourapproach consistently outperforms the unimodal models. This advantage arises from \ourapproach's integration of additional modal information.
    This demonstrates \ourapproach's ability to effectively decouple modality parameters, enabling it to handle inputs from different modalities more efficiently, and highlights its strength in enhancing multimodal understanding.
    }
    
    \subsection{Detailed Results}
    \label{app:results}
    In this section, we present detailed results from the multi-modality retention and mitigating catastrophic forgetting experiments. 
     The results of various baselines for seven vision tasks are shown in Table~\ref{table2_detail_I}, two point cloud tasks in Table~\ref{table2_detail_P}, three audio tasks and two video tasks in Table~\ref{table2_detail_AV}, three multimodal tasks in Table~\ref{table3_detail_multimodal}, and the last two new tasks in Table~\ref{table3_detail_new}.

\input{tables/table2_detail_AV}
\input{tables/table3_detail_multimodal}
\input{tables/table3_detail_new}

  \section{Qualitative Results}
  We provide qualitative results in Figure~\ref{fig:exam2}.
  These results demonstrate the capability of the merged MLLM constructed by our \ourapproach approach to understand and reason with multimodal inputs.
   \begin{figure*}[t]
		\centering
		\includegraphics[width=0.95\linewidth]{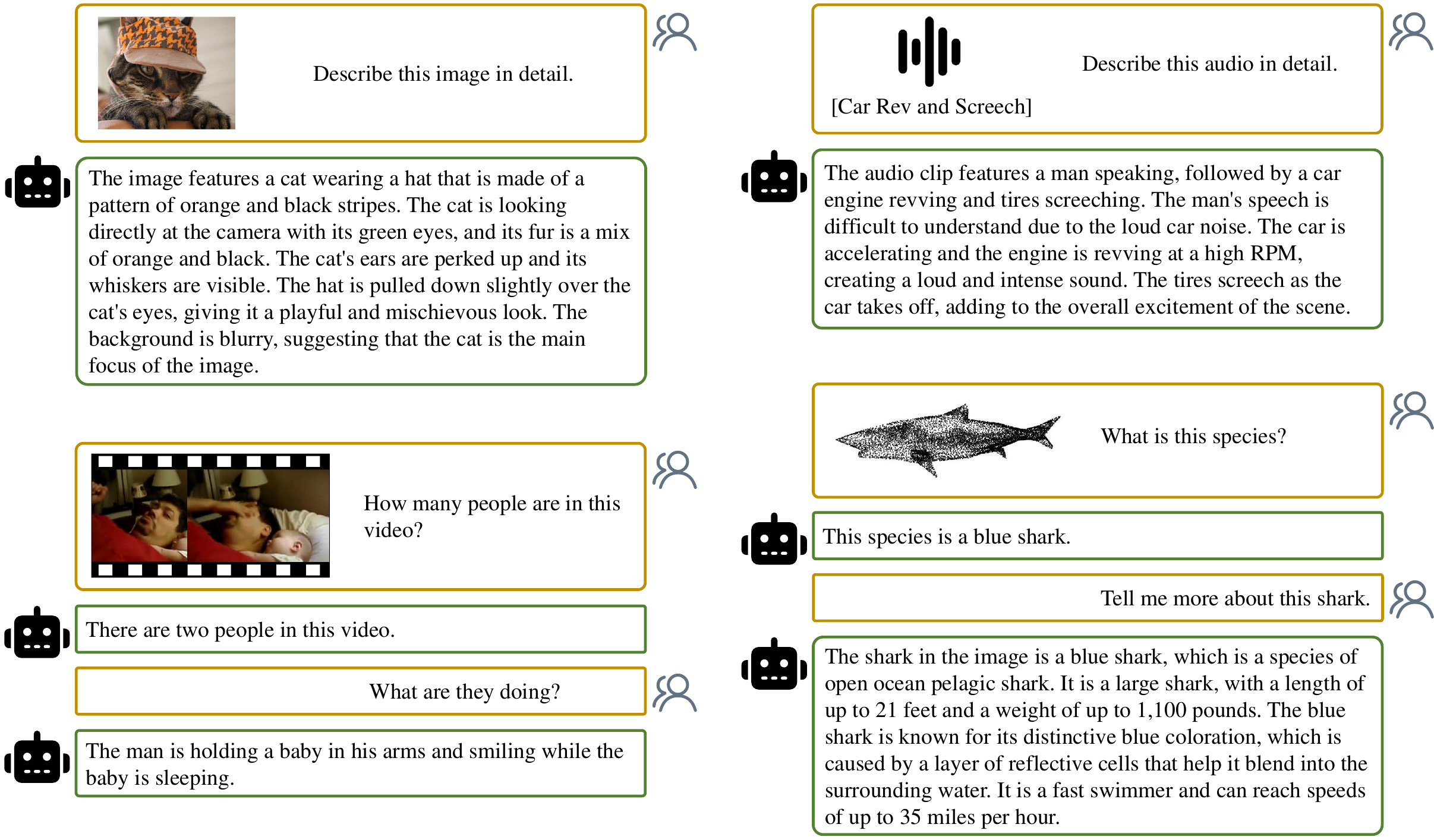}
        \vspace{3pt}
	\end{figure*}

 \begin{figure*}[t]
		\centering
		\includegraphics[width=0.95\linewidth]{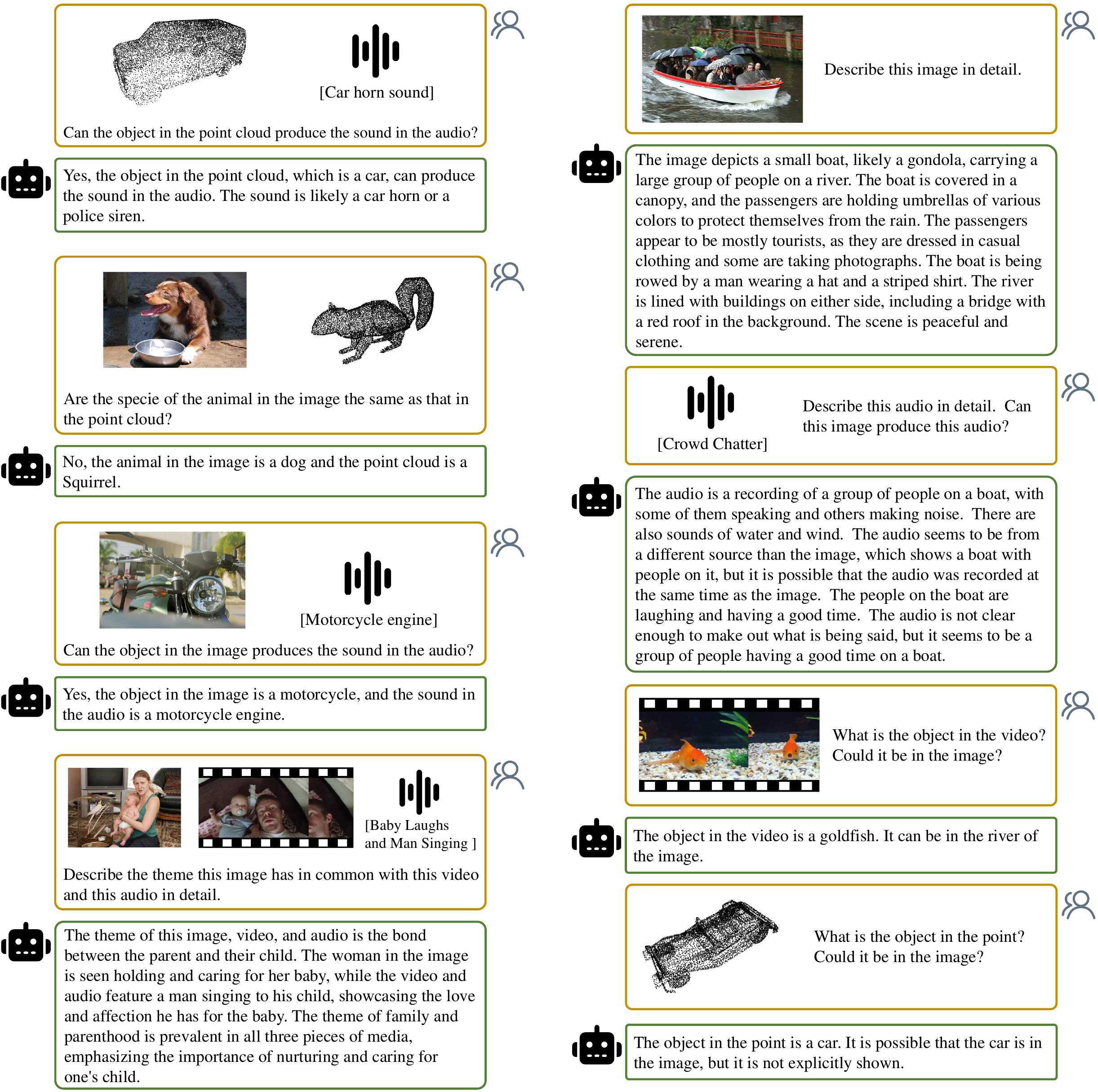}
\vspace{-5pt}
		\caption{   Qualitative results.
		}
        \vspace{-7pt}
		\label{fig:exam2}
	\end{figure*}

    \section{Prompt for Evaluation}
     We present the evaluation prompts for each benchmark in Table~\ref{table:prompts}. To denote the inputs for various modalities, we use ``<image>'', ``<audio>'', ``<video>'', and ``<point>'' to represent vision, audio, video, and point cloud modalities, respectively.

     \input{tables/table_prompts}
     
\end{document}

%% file: math_commands.tex
\usepackage{amsmath,amsfonts,bm}

\def\eqref#1{equation~\ref{#1}}

\def\1{\bm{1}}

\DeclareMathAlphabet{\mathsfit}{\encodingdefault}{\sfdefault}{m}{sl}
\SetMathAlphabet{\mathsfit}{bold}{\encodingdefault}{\sfdefault}{bx}{n}

\DeclareMathOperator*{\argmin}{arg\,min}

%% file: tables/table1.tex
\begin{table*}[t]
\centering
\renewcommand{\arraystretch}{1.25}
\captionsetup{type=table}

\resizebox{1.0\linewidth}{!}{  
\begin{tabular}{l|c|cccc|ccccc|c}
\bottomrule
\rowcolor{mygray}
Task ($\rightarrow$) &  \textbf{ModelNet40} & \multicolumn{4}{c|}{ \textbf{MUSCI-AVQA} } & \multicolumn{5}{c|}{\centering \textbf{MCUB}} & \\ 
\rowcolor{mygray}
Method ($\downarrow$) & PI-T & IA-T & VI-T &VA-T &AVI-T & AVI-T & AVP-T & AIP-T & VIP-T & AVIP-T  & \multirow{-2}{*}{Avg.}\\

\hline
\textit{\small --Training-based Multimodal Baselines} & & & & & & & & & & & \\  

{ImageBind-LLM\citep{32}} &{39.86}	&{36.54}	&{38.76}	&{39.72} &{38.16}	&{35.20}	&{31.40}	&{33.40}	&{31.80}	&{32.93}		&{35.51} \\

{X-InstructBLIP\pub{ECCV24} \citep{20}} &{57.93}	&{40.71}	&{41.23}	&{48.34} &{47.39}	&{41.40}	&{25.20}	&{21.20}	&{29.40}	&{27.94}		&{37.04} \\
\hline
\textit{\small --Training-free Model Merging Methods} & & & & & & & & & & &\\  

NaiveMC\pub{ACL24} \citep{18} &60.53	&39.31	&47.65	&47.40 &49.64 &53.64	&56.28	&60.53	&54.60	&59.16		&53.23 \\

TA\pub{ICLR23} \citep{21} &62.04	&40.22	&47.97	&46.70 &49.93  & 53.44	&56.28	&63.36	&55.40	&59.72		&53.90 \\

TIES\pub{NeurIPS23} \citep{24} &61.74	&43.27	&49.27	&48.60  &51.19 & 53.64	&55.47	&61.74	&54.60	&58.55		&54.10 \\

PCB-Merging\pub{NeurIPS24} \citep{84} &62.15	&\underline{44.32}	&\underline{50.24}	&\underline{49.67}  &\underline{51.54}  &\underline{54.54}	&{56.68}	&63.97	&\underline{55.60}	&\underline{60.48}		&\underline{54.92} \\

NaiveMC (w/ DARE\pub{ICML24} \citep{43})  &60.32	&39.78	&47.98	&47.67 &49.89 & 53.64	&\underline{56.68}	&60.73	&54.80	&59.53	&53.46
 \\

TA (w/ DARE) &\textbf{62.75 \textcolor{color2}}	&40.46	&47.98	&46.92 &50.27 & 54.25	&56.48	&\underline{64.17}	&{55.40}	&{60.08}	&54.27 \\

TIES (w/ DARE) &61.96	&{43.78}	&{49.54}	&{48.98}  &{51.36} &{54.25}	&{55.87}	&62.55	&55.20	&59.06	&{54.57} \\

\cellcolor[HTML]{FDF5F1}\textbf{\ourapproach (ours)}  
& 
 \cellcolor[HTML]{FDF5F1}\underline{62.15}  & \cellcolor[HTML]{FDF5F1}\textbf{47.25} & \cellcolor[HTML]{FDF5F1}\textbf{51.27} & \cellcolor[HTML]{FDF5F1}\textbf{51.77}
 & \cellcolor[HTML]{FDF5F1}\textbf{53.54}
 & \cellcolor[HTML]{FDF5F1}\textbf{56.48} & \cellcolor[HTML]{FDF5F1}\textbf{59.31 } & \cellcolor[HTML]{FDF5F1}\textbf{65.59 } & \cellcolor[HTML]{FDF5F1}\textbf{56.00 } & \cellcolor[HTML]{FDF5F1}\textbf{61.63 } & \cellcolor[HTML]{FDF5F1}\textbf{56.82}\\ 
\hline

\end{tabular}

}
\vspace{-4pt}
\caption{\label{tab:main} Accuracy (\%) on multimodal tasks with various combinations of video (V), image (I), audio (A), point cloud (P), and text (T) inputs. 
Optimal results are in bold, while sub-optimal results are underlined.}
\label{table1}
\end{table*}

%% file: tables/table2.tex
  \begin{table*}[t]
  \renewcommand{\arraystretch}{1.48}

  \centering
  \resizebox{1\linewidth}{!}{
  \begin{tabular}{l|cccc|c}

      \bottomrule
      
\rowcolor{mygray}
Task ($\rightarrow$) & { \textbf{2 Point Tasks}}   &  \textbf{3 Audio Tasks} &  \textbf{2 Video Tasks} & { \textbf{7 Image Tasks} } &{ \textbf{Trimmed Avg.} } \\ 

\rowcolor{mygray}
Method ($\downarrow$)  & Score (\%) / Acc. (\%) & Score (\%) / Acc. (\%) & Acc. (\%) & Acc. (\%) & Score (\%) / Acc. (\%) \\
      
      \hline
     \cellcolor[HTML]{F5F5F5}{Original MLLMs (Zero-shot)}  &\cellcolor[HTML]{F5F5F5}23.15 / 21.27 & \cellcolor[HTML]{F5F5F5}25.30 / 24.71& \cellcolor[HTML]{F5F5F5}39.79 & \cellcolor[HTML]{F5F5F5}62.23 &\cellcolor[HTML]{F5F5F5}24.23 / 51.01  \\
  {NaiveMC} \pub{ACL2024} \citep{18} & \foo{22.65}{97.8} / \foo{20.49}{96.3} &\foo{24.59}{97.2} / \foo{30.65}{124.8}  & \foo{36.92}{93.0} &\foo{52.56}{83.6} & \foo{23.62}{97.5} / \foo{44.59}{88.3}    \\

  {TA} \pub{ICLR23} \citep{21} & \foo{22.96}{99.2} / \foo{21.02}{98.8} & \foo{24.68}{97.5} / \foo{31.88}{129.8} & \foo{37.57}{94.5} & \foo{54.89}{87.5}& \foo{23.82}{98.3}  / \foo{46.23}{91.0}       \\
  
  {TIES} \pub{NeurIPS23} \citep{24} & \foo{22.82}{98.6} / \foo{20.83}{97.9} & \foo{24.79}{98.0} / \foo{32.15}{130.9} & \foo{37.81}{95.1} & \foo{54.10}{86.2}& \foo{23.80}{98.3}  / \foo{45.96}{90.6}       \\

  {PCB-Merging} \pub{NeurIPS24} \citep{84} & \foo{23.00}{99.4} / \foo{21.16}{99.5} &\underline{ \foo{25.03}{98.9} / \foo{33.41}{135.2}} & \underline{\foo{38.47}{96.7}} &\underline{ \foo{56.02}{90.0}}&\underline{ \foo{24.02}{99.1}  / \foo{47.24}{92.6}}       \\

{NaiveMC (w/ DARE\pub{ICML2024} \citep{43})} & \foo{22.83}{98.6} / \foo{20.77}{97.6} &\foo{24.72}{97.7} / \foo{31.62}{128.8}  & \foo{37.63}{94.4} &\foo{53.61}{85.3} & \foo{23.78}{98.1} / \foo{45.62}{89.8}    \\

  {TA (w/ DARE)}  & \underline{\foo{23.04}{99.5} / \foo{21.25}{99.9}} & \foo{24.82}{98.1} / \foo{32.44}{132.0} & \foo{37.52}{94.4} & {\foo{55.47}{88.4}}& {\foo{23.95}{98.8}  / \foo{46.50}{91.4}}       \\
  
  {TIES (w/ DARE)}  & \foo{22.76}{98.3} / \foo{20.98}{98.6} & {\foo{24.92}{98.5} / \foo{33.02}{134.4}} & {\foo{38.00}{95.6}} & \foo{54.73}{87.2}& \foo{23.84}{98.4}  / \foo{46.37}{91.4}       \\
  
  {\cellcolor[HTML]{FDF5F1}\textbf{\ourapproach (ours)}}  & {\cellcolor[HTML]{FDF5F1}\textbf{\foo{23.14}{99.9} / \foo{22.49}{105.7}}} & {\cellcolor[HTML]{FDF5F1}\textbf{\foo{25.20}{99.6} / \foo{38.51}{155.6}}}
  &{\cellcolor[HTML]{FDF5F1}\textbf{\foo{39.28}{98.5}}}
  & {\cellcolor[HTML]{FDF5F1}\textbf{\foo{62.40}{100.3}}}
& {\cellcolor[HTML]{FDF5F1}\textbf{\foo{24.17}{99.8}} /   \textbf{\foo{50.84}{99.4}}}       \\
  \hline

  \end{tabular}
  }
  \vspace{-4pt}
  \caption{Results of multi-modality retention experiments.
    	The performance retention is shown in parentheses.
    	``Trimmed Avg.'' represents the average result obtained after excluding three point or audio classification tasks. }
          \label{table2}
  \vspace{-6pt}
  \end{table*}

%% file: tables/table3.tex
  \begin{table*}[t]
    
  \centering
    \renewcommand{\arraystretch}{1.25}
  \resizebox{1\linewidth}{!}{
  \begin{tabular}{l|ccccc|cc}

      \bottomrule
      

\rowcolor{mygray}
& \multicolumn{5}{c|}{\centering \textbf{Previous Tasks}} & \multicolumn{2}{c}{\textbf{New Tasks}} \\ 
  
\rowcolor{mygray}
\multirow{-2}{*}{Task ($\rightarrow$)} & { \textcolor[rgb]{0.27, 0.35, 0.760} {2 Point tasks}}   &\textcolor[rgb]{0.27, 0.35, 0.760}   {3 Audio tasks} & \textcolor[rgb]{0.27, 0.35, 0.760}  {2 Video tasks} & {\textcolor[rgb]{0.27, 0.35, 0.760}  {7 Image tasks} } &{\textcolor[rgb]{0.27, 0.35, 0.760} {3 Multimodal tasks} } &{\textcolor[rgb]{0.773,0.353,0.067}{Clotho-AQA} } &{ \textcolor[rgb]{0.773,0.353,0.067}{Flickr30k}}\\ 

\rowcolor{mygray}
Baseline ($\downarrow$)  & \textcolor[rgb]{0.27, 0.35, 0.760}{Score / Acc.} & \textcolor[rgb]{0.27, 0.35, 0.760}{Score  / Acc.} & \textcolor[rgb]{0.27, 0.35, 0.760}{Acc.}  & \textcolor[rgb]{0.27, 0.35, 0.760}{Acc.}  &  \textcolor[rgb]{0.27, 0.35, 0.760}{Acc.} & \textcolor[rgb]{0.773,0.353,0.067} {Acc.} &  \textcolor[rgb]{0.773,0.353,0.067}{Score}\\

\hline
      \cellcolor[HTML]{F5F5F5}{Original MLLMs}  &\cellcolor[HTML]{F5F5F5}23.15 / 21.27 & \cellcolor[HTML]{F5F5F5} 25.30 / 24.71&\cellcolor[HTML]{F5F5F5}39.79& \cellcolor[HTML]{F5F5F5}62.23&\cellcolor[HTML]{F5F5F5}- &\cellcolor[HTML]{F5F5F5}49.40 & \cellcolor[HTML]{F5F5F5}51.26 \\

    {Fine-tune on Clotho-AQA}  & - &19.82 / 12.31 \textcolor{red}{($\downarrow$)}  &- & - &- & 57.80 \textcolor{color_green}{($\uparrow$)} & -     \\
    
    {Fine-tune on Flickr30k}  & - & - &- & 57.25 \textcolor{red}{($\downarrow$)} &- & -& 57.71 \textcolor{color_green}{($\uparrow$)}     \\
      \hline
      \cellcolor[HTML]{F5F5F5}{\ourapproach}  &\cellcolor[HTML]{F5F5F5}23.14 / 22.49 &\cellcolor[HTML]{F5F5F5}25.20 / 38.51&\cellcolor[HTML]{F5F5F5}39.28&\cellcolor[HTML]{F5F5F5}62.40&\cellcolor[HTML]{F5F5F5}56.82 &\cellcolor[HTML]{F5F5F5}{49.28} & \cellcolor[HTML]{F5F5F5}{51.00} \\
    
  
  \cellcolor[HTML]{FDF5F1}{\textbf{\ourapproach-Clotho-AQA}}  & {\cellcolor[HTML]{FDF5F1}{22.95 / 21.87}} & {\cellcolor[HTML]{FDF5F1}{{25.12 / 38.23} \textcolor{blue}{($\sim$)}}}& \cellcolor[HTML]{FDF5F1}{39.17}&\cellcolor[HTML]{FDF5F1}62.20 & \cellcolor[HTML]{FDF5F1}{56.53}& \cellcolor[HTML]{FDF5F1}{{57.71} \textcolor{color_green}{($\uparrow$)}} &\cellcolor[HTML]{FDF5F1}50.94        \\
  
  \cellcolor[HTML]{FDF5F1}{\textbf{\ourapproach-Flickr30k}}  & \cellcolor[HTML]{FDF5F1}{23.05 / 22.03} & \cellcolor[HTML]{FDF5F1}{24.96 / 37.68}& \cellcolor[HTML]{FDF5F1}{38.90}&\cellcolor[HTML]{FDF5F1}{62.27} \textcolor{blue}{($\sim$)} & \cellcolor[HTML]{FDF5F1}{56.44}&\cellcolor[HTML]{FDF5F1} 48.94 &\cellcolor[HTML]{FDF5F1}{{57.08} \textcolor{color_green}{($\uparrow$)}}          \\
\hline
  
  \cellcolor[HTML]{F8E8E3}{\textbf{\ourapproach-Clotho-AQA+Flickr30k}}  &\cellcolor[HTML]{F8E8E3}{22.82 / 21.56} 
  & \cellcolor[HTML]{F8E8E3}{{24.88 / 37.69} \textcolor{blue}{($\sim$)}}
  & \cellcolor[HTML]{F8E8E3}{38.53}
  &\cellcolor[HTML]{F8E8E3}{61.94} \textcolor{blue}{($\sim$)} 
  & \cellcolor[HTML]{F8E8E3}{55.89}
  & \cellcolor[HTML]{F8E8E3}{{57.52} \textcolor{color_green}{($\uparrow$)}}
  &\cellcolor[HTML]{F8E8E3}{{56.72} \textcolor{color_green}{($\uparrow$)}}         \\
  \hline

  \end{tabular}
  }
  \vspace{-2.5pt}
  \caption{Results on previous and new tasks in both single-task and \textbf{cross-modal} multi-task scenario. 
    \ourapproach-xx refers to merging the MLLM fine-tuned on the new task xx into \ourapproach. 
    \ourapproach-Clotho-AQA+Flickr30k denotes the merging of both the audio LLM fine-tuned on Clotho-AQA and the vision LLM fine-tuned on Flickr30k into \ourapproach.
    }
    \vspace{-2pt}
  \label{table3}

  \end{table*}

%% file: tables/table5.tex
\setlength{\intextsep}{3pt}
\setlength{\columnsep}{10pt}
\begin{table}

\renewcommand{\arraystretch}{1.1}
\centering
\captionsetup{type=table}

\resizebox{\linewidth}{!}{  
\begin{tabular}{lcc}
\bottomrule
\rowcolor{mygray}
\textbf{} & \textbf{Expansion} & \textbf{Retention} \\
\rowcolor{mygray}
\multirow{-2}{*}{\textbf{Method}} & \textbf{ACC.} & \textbf{Score (\%) / ACC. (\%)} \\
\hline
\textbf{\ourapproach} & \textbf{56.82} & \textbf{\foo{24.17}{99.8} / \foo{50.84}{99.4}}  \\
\hline
\; $-$ \small {Directional Congruence} & 7.20 & \foo{10.05}{41.6} / \foo{8.34}{16.7} \\
\; $-$ \small {Dominant Significance} & 33.87 & \foo{14.71}{60.5} / \foo{28.93}{57.1} \\
\; $-$ \small {Scaling Factor} $\lambda$ & 54.02 & \foo{23.14}{95.6} / \foo{47.78}{93.9} \\
\bottomrule
\end{tabular}
}
  \vspace{-5pt}
\caption{ Ablation study on parameter decoupling steps.
}
  \vspace{-10.7pt}
\label{table5}
\end{table}

%% file: tables/table4.tex
  \begin{table*}[t]
\renewcommand{\arraystretch}{1.1}
\captionsetup{type=table}

\centering
\resizebox{\linewidth}{!}{  
\begin{tabular}{l|cc|cc|c}
\bottomrule
\rowcolor{mygray}
\textbf{} & \multicolumn{2}{c|}{\centering \textbf{One New Task}} & \multicolumn{2}{c|}{\centering \textbf{Two New Tasks}} & \\
\rowcolor{mygray}
\multirow{-2}{*}{\textbf{Method}} & {Previous tasks} &New task & {Previous tasks} &New tasks &\multirow{-2}{*}{\textbf{Storage}}  \\
\hline
{Model Tailor}\pub{ICML24} \citep{49} &{96.47 $\%$} & {91.69 $\%$} & {99.28 $\%$} & {87.50 $\%$}  &32$(P + P')$ \\
\hline

\textbf{\ourapproach (ours)} & \textbf{99.86 \%} & \textbf{99.67 \%} & \textbf{99.63 \%} & \textbf{99.42 \%} &64$P$ + 32$P'$ + $NP$ \\

\bottomrule
\end{tabular}
}
\caption{ Performance retention $\&$ Storage vs. Mitigating MLLMs' catastrophic forgetting methods in the \textbf{same modality}. 
Let $N$, $P$, and $P'$  represent the number of new tasks, the total LLM parameters, and the modality-specific component parameters, assuming each float parameter occupies 32 bits.}
\label{table4}
  \end{table*}

%% file: tables/table_mllms.tex
\begin{table*}[ht]
\centering
\scriptsize
\renewcommand{\arraystretch}{1.25}
\begin{tabular}{
  >{\raggedright\arraybackslash}p{0.4in}
  >{\raggedright\arraybackslash}p{0.8in}
  >{\raggedright\arraybackslash}p{0.4in}
  >{\raggedright\arraybackslash}p{0.9in}
  >{\raggedright\arraybackslash}p{1.2in}
  >{\raggedright\arraybackslash}p{0.8in}
}
\bottomrule
\rowcolor{mygray}
Modality & Modality Encoder & Connector & Alignment Data & Fine-tuneing Data &Referenced Work \\ \hline

Image & CLIP-ViT-L-336px ~\citep{72}  & MLP & LCS 558K~\citep{53} & LLaVA-mixed 665K~\citep{53} &  LLaVA-1.5~\citep{52}\\

\midrule
Audio & BEATs-Iter3+~\citep{77} & Q-Former & WaveCaps 400K~\citep{73} & OpenAQA filtered 350K~\citep{74} &X-InstructBLIP ~\citep{20} \\

\midrule
Video & LanguageBind~\citep{76} & MLP & LCS 558K,\quad\quad\quad Valley 702K~\citep{75} & Video-ChatGPT 100K~\citep{11}, LLaVA-mixed sampled 140K & Video-LLaVA~\citep{10} \\
\midrule

Point Cloud & Point Encoder~\citep{53} & MLP & PointLLM brief description 660K~\citep{53}  & Point complex instruction 70K ~\citep{53} &  PointLLM~\citep{53} \\ \bottomrule
\end{tabular}
\caption{Training data and components of MLLMs for different modalities.}
\label{table:mllms}
\end{table*}

%% file: tables/table_hyper.tex
\begin{table*}[ht]
\centering
\small
\renewcommand{\arraystretch}{1.2}
\begin{tabular}{llcccc}
\bottomrule
\rowcolor{mygray}
Stage & Hyperparameter & Image & Audio & Video & Point Cloud  \\ \hline

\multirow{4}{*}{Alignment-State} 
& Batch size & 256 & 256 & 256 & 128 \\
& LR & 1e-3 & 1e-3 & 1e-3 & 2e-3 \\
& LR Schedule & \multicolumn{4}{c}{cosine decay} \\
& Warmup Ratio & \multicolumn{4}{c}{0.03} \\
& Epoch & 1 & 1 & 1 & 3 \\
\midrule

\multirow{4}{*}{Fine-tuning-Stage} 
& Batch size & 128 & 64 & 128 & 64 \\
& LR & 2e-5 & 1e-5 & 2e-5 & 2e-5 \\
& LR Schedule & \multicolumn{4}{c}{cosine decay} \\
& Warmup Ratio & \multicolumn{4}{c}{0.03} \\
& Epoch & 1 & 3 & 1 & 3 \\
\bottomrule
\end{tabular}
\caption{Hyperparameters of different MLLMs.}
\label{table:hyper}
\end{table*}

%% file: tables/l1l2.tex
\setlength{\intextsep}{3pt}
\setlength{\columnsep}{10pt}
\begin{table}
\renewcommand{\arraystretch}{1.1}
\centering
\captionsetup{type=table}

\resizebox{\linewidth}{!}{  
\begin{tabular}{lcc}
\bottomrule
\rowcolor{mygray}
\textbf{} & \textbf{Expansion} & \textbf{Retention} \\
\rowcolor{mygray}
\multirow{-2}{*}{} & \textbf{ACC.} & \textbf{Score / ACC.} \\
\hline
\textbf{\ourapproach (Manhattan)} & {56.82} & {24.17} / {50.84}  \\
\textbf{\ourapproach (Euclidean)} & {56.05} & {23.89} / {50.41}  \\

\bottomrule
\end{tabular}
}
\caption{ {Results of \ourapproach with Manhattan distance or Euclidean distance}
}
\vspace{-10pt}
\label{l1l2}
\end{table}

%% file: tables/table_tvs.tex
\setlength{\intextsep}{3pt}
\setlength{\columnsep}{10pt}
\begin{table}
\renewcommand{\arraystretch}{1.1}
\centering
\captionsetup{type=table}

\resizebox{\linewidth}{!}{  
\begin{tabular}{lcc}
\bottomrule
\rowcolor{mygray}
\textbf{} & \textbf{Directional Alignment} & \textbf{Average Magnitude}  \\

\hline
{Vision} &69.20\%  &5e-4 \\
{Audio} &50.62\%  &8e-5\\
{Video} &57.58\%  &2e-4 \\
{Point} &70.09\%  &5e-4\\

\bottomrule
\end{tabular}
}
\caption{ {Percentage of parameters whose directions align with those in the merged task vector and the average magnitude of the parameters across the task vectors of the four modalities}
}
\label{table_tvs}
\end{table}

%% file: tables/table_lora.tex
  \begin{table*}[t]

  \centering
    \renewcommand{\arraystretch}{1.2}
  \resizebox{1\linewidth}{!}{
  \begin{tabular}{l|cccccccc|c}

      \bottomrule

& \multicolumn{8}{c|}{\centering \textbf{7 Original Image Tasks}}  & \multicolumn{1}{c}{\centering \textbf{New Tasks}}\\

\multirow{-2}{*}{Task ($\rightarrow$)} & { \textcolor[rgb]{0.27, 0.35, 0.760} {VQAv2}}   &\textcolor[rgb]{0.27, 0.35, 0.760}   {GQA} & \textcolor[rgb]{0.27, 0.35, 0.760}  {TextVQA} & {\textcolor[rgb]{0.27, 0.35, 0.760}  {VizWiz} } &{\textcolor[rgb]{0.27, 0.35, 0.760} {ScienceQA} } &{\textcolor[rgb]{0.27, 0.35, 0.760}{POPE} } &{ \textcolor[rgb]{0.27, 0.35, 0.760}{OK-VQA}} &{ \textcolor[rgb]{0.27, 0.35, 0.760}{Avg.}}  &{ \textcolor[rgb]{0.773,0.353,0.067}{Flickr30k}} \\

Method ($\downarrow$)  &\textcolor[rgb]{0.27, 0.35, 0.760} {Acc.} &\textcolor[rgb]{0.27, 0.35, 0.760} {Acc.} &\textcolor[rgb]{0.27, 0.35, 0.760} {Acc.}  &\textcolor[rgb]{0.27, 0.35, 0.760} {Acc.}  & \textcolor[rgb]{0.27, 0.35, 0.760} {Acc.} & \textcolor[rgb]{0.27, 0.35, 0.760}{Acc.} & \textcolor[rgb]{0.27, 0.35, 0.760} {Acc.} & \textcolor[rgb]{0.27, 0.35, 0.760} {Acc.} &  \textcolor[rgb]{0.773,0.353,0.067}{Score} \\

\hline
      \cellcolor[HTML]{F5F5F5}{Original MLLMs}  &\cellcolor[HTML]{F5F5F5}78.11 & \cellcolor[HTML]{F5F5F5}61.52&\cellcolor[HTML]{F5F5F5}55.89& \cellcolor[HTML]{F5F5F5}51.51&\cellcolor[HTML]{F5F5F5}71.12 &\cellcolor[HTML]{F5F5F5}86.17 & \cellcolor[HTML]{F5F5F5}31.33 & \cellcolor[HTML]{F5F5F5}62.23 & \cellcolor[HTML]{F5F5F5}51.26\\
    Fine-tune on Flickr30k &72.27 &54.19 &46.10 &52.88 &70.22 &76.28 &28.31 &57.25 &57.71 \\
    Lora &75.72 &58.24 &52.87 &52.64 &70.63 &85.08 &29.21 &60.63 &54.85 \\
  {\cellcolor[HTML]{FDF5F1}\textbf{\ourapproach-Flickr30k (ours)}}  &\cellcolor[HTML]{FDF5F1}77.75 &\cellcolor[HTML]{FDF5F1}61.43 &\cellcolor[HTML]{FDF5F1}55.41 &\cellcolor[HTML]{FDF5F1}52.72 &\cellcolor[HTML]{FDF5F1}71.75 &\cellcolor[HTML]{FDF5F1}85.72 &\cellcolor[HTML]{FDF5F1}31.07 &\cellcolor[HTML]{FDF5F1}62.27
  &\cellcolor[HTML]{FDF5F1}57.08\\

\hline

  \end{tabular}
  }
  \caption{{The results of \ourapproach and LoRA fine-tuning on original vision LLM for Flickr30k.}
    }
    
  \label{table_lora}

  \end{table*}

%% file: tables/table_unimodal.tex
\begin{table}[t]
\centering
\renewcommand{\arraystretch}{1.33}
\captionsetup{type=table}

\small

\resizebox{1.0\linewidth}{!}{  
\begin{tabular}{l|cc}
\bottomrule
\rowcolor{mygray}
Task ($\rightarrow$) & & \\ 
\rowcolor{mygray}
Model ($\downarrow$)& \small \multirow{-2}{*}{ModelNet40} & \small \multirow{-2}{*}{MUSCI-AVQA} \\

\hline
{Vision MLLM} &51.94 &44.06\\ 
{Audio MLLM} &- &30.63 \\
{Video MLLM} &- &47.72 \\
{Point MLLM} &21.27 &- \\
\cellcolor[HTML]{FDF5F1}\textbf{{\ourapproach (ours)}} & \cellcolor[HTML]{FDF5F1}62.15 & \cellcolor[HTML]{FDF5F1}53.54 \\

\hline

\end{tabular}

}
\caption{{Accuracy (\%) results of four original unimodal models on the multimodal tasks.}}
\label{table_unimodal}
\end{table}

%% file: tables/table2_detail_I.tex
  \begin{table*}[t]
  \centering
    \renewcommand{\arraystretch}{1.2}
  \resizebox{1\linewidth}{!}{
  \begin{tabular}{l|ccccccc}

      \bottomrule

& \multicolumn{7}{c}{\centering \textbf{7 Image Tasks}} \\

\multirow{-2}{*}{Task ($\rightarrow$)} & { \textcolor[rgb]{0.27, 0.35, 0.760} {VQAv2}}   &\textcolor[rgb]{0.27, 0.35, 0.760}   {GQA} & \textcolor[rgb]{0.27, 0.35, 0.760}  {TextVQA} & {\textcolor[rgb]{0.27, 0.35, 0.760}  {VizWiz} } &{\textcolor[rgb]{0.27, 0.35, 0.760} {ScienceQA} } &{\textcolor[rgb]{0.27, 0.35, 0.760}{POPE} } &{ \textcolor[rgb]{0.27, 0.35, 0.760}{OK-VQA}}\\

Method ($\downarrow$)  & Acc. & Acc. & Acc.  & Acc.  &  Acc. & Acc. &  Acc. \\

\hline
      \cellcolor[HTML]{F5F5F5}{Original MLLMs}  &\cellcolor[HTML]{F5F5F5}78.11 & \cellcolor[HTML]{F5F5F5}61.52&\cellcolor[HTML]{F5F5F5}55.89& \cellcolor[HTML]{F5F5F5}51.51&\cellcolor[HTML]{F5F5F5}71.12 &\cellcolor[HTML]{F5F5F5}86.17 & \cellcolor[HTML]{F5F5F5}31.33 \\
      
  {\cellcolor[HTML]{FDF5F1}\textbf{\ourapproach (ours)}}  &\cellcolor[HTML]{FDF5F1}77.95 &\cellcolor[HTML]{FDF5F1}61.85 &\cellcolor[HTML]{FDF5F1}55.74 &\cellcolor[HTML]{FDF5F1}52.26 &\cellcolor[HTML]{FDF5F1}71.16 &\cellcolor[HTML]{FDF5F1}86.58 &\cellcolor[HTML]{FDF5F1}31.27       \\
      
\hline
\textit{\small --Multi-Modality Retention} \\
    {NaiveMC} \pub{ACL2024} \citep{18} & 59.73 &45.83  & 42.29 &47.87 & 68.52 &79.41 &24.28    \\
    
    {TA} \pub{ICLR23} \citep{21} &62.71 &48.86 &45.20 &49.47 & 70.04 & 82.38 & 25.56       \\
  
  {TIES} \pub{NeurIPS23} \citep{24} & 61.78 &48.23 &44.60 &48.67 &69.05 &81.21 & 25.13       \\

{NaiveMC (w/ DARE\pub{ICML2024} \citep{43})}  & 60.91 &46.62 &42.88 &49.04 &70.09 &81.08 &24.62    \\

  {TA (w/ DARE)}  & 63.65 &49.25 &45.74 &49.82 &70.87 &83.12 &25.82       \\
  
  {TIES (w/ DARE)}  & 62.54 &48.73 &45.38 &49.15 &69.78 &82.17 &25.39       \\
  
  
  \hline
  \textit{\small --Mitigating Catastrophic Forgetting} \\
  
  {Fine-tune on Flickr30k}  & 72.27 &54.19 &46.10 &52.88 &70.22 &76.78 &28.31 \\
  
   {\cellcolor[HTML]{FDF5F1}\textbf{\ourapproach-Clotho-AQA}}  &\cellcolor[HTML]{FDF5F1}77.87 &\cellcolor[HTML]{FDF5F1}61.59 &\cellcolor[HTML]{FDF5F1}55.51 &\cellcolor[HTML]{FDF5F1}51.88 &\cellcolor[HTML]{FDF5F1}71.16 &\cellcolor[HTML]{FDF5F1}86.24 &\cellcolor[HTML]{FDF5F1}31.14       \\

   {\cellcolor[HTML]{FDF5F1}\textbf{\ourapproach-Flickr30k}}  &\cellcolor[HTML]{FDF5F1}77.75 &\cellcolor[HTML]{FDF5F1}61.43 &\cellcolor[HTML]{FDF5F1}55.41 &\cellcolor[HTML]{FDF5F1}52.72 &\cellcolor[HTML]{FDF5F1}71.75 &\cellcolor[HTML]{FDF5F1}85.72 &\cellcolor[HTML]{FDF5F1}31.07       \\

   {\cellcolor[HTML]{FDF5F1}\textbf{\ourapproach-Clotho-AQA+Flickr30k}}  &\cellcolor[HTML]{FDF5F1}77.32 &\cellcolor[HTML]{FDF5F1}61.33 &\cellcolor[HTML]{FDF5F1}55.23 &\cellcolor[HTML]{FDF5F1}52.33 &\cellcolor[HTML]{FDF5F1}71.02 &\cellcolor[HTML]{FDF5F1}85.43 &\cellcolor[HTML]{FDF5F1}30.94       \\

\hline

  \end{tabular}
  }
    \caption{Results for each method on seven image tasks. All tasks are Question-Answering tasks.
    }
    \vspace{-10pt}
  \label{table2_detail_I}
  \end{table*}

%% file: tables/table2_detail_P.tex
  \begin{table*}[t]

  \centering
    \renewcommand{\arraystretch}{1.2}
  \resizebox{1\linewidth}{!}{
  \begin{tabular}{l|c|ccccc}

      \bottomrule

& \multicolumn{6}{c}{\centering \textbf{2 Point Tasks}} \\ 

\cline{2-7}

\multirow{-2}{*}{Task ($\rightarrow$)} & { \textcolor[rgb]{0.27, 0.35, 0.760} {ModelNet40}}   &\multicolumn{5}{c}{\centering\textcolor[rgb]{0.27, 0.35, 0.760} { Objavers-captioning}}  \\

Method ($\downarrow$)  & Acc. & BLEU-1 & ROUGE-L  & METEOR  &  Sentence-BERT & SimCSE \\   

\hline
      \cellcolor[HTML]{F5F5F5}{Original MLLMs}  
      &\cellcolor[HTML]{F5F5F5}21.27&\cellcolor[HTML]{F5F5F5}4.73 & \cellcolor[HTML]{F5F5F5}8.51&\cellcolor[HTML]{F5F5F5}12.02& \cellcolor[HTML]{F5F5F5}44.18&\cellcolor[HTML]{F5F5F5}46.31 \\

  {\cellcolor[HTML]{FDF5F1}\textbf{\ourapproach (ours)}} 
  &\cellcolor[HTML]{FDF5F1}22.49
  &\cellcolor[HTML]{FDF5F1}5.06 &\cellcolor[HTML]{FDF5F1}8.53 &\cellcolor[HTML]{FDF5F1}11.90 &\cellcolor[HTML]{FDF5F1}43.72 &\cellcolor[HTML]{FDF5F1}46.51     \\
\hline
\textit{\small –Multi-Modality Retention} & \\
    {NaiveMC} \pub{ACL2024} \citep{18} &20.49 & 4.43 &8.24 &11.37 &43.22 &45.97    \\
    
    {TA} \pub{ICLR23} \citep{21} &21.02 &4.69 &8.46 &11.73 &43.55 &46.38       \\
  
  {TIES} \pub{NeurIPS23} \citep{24} &20.83 & 4.55 &8.39 &11.60 &43.29 &46.27       \\

{NaiveMC (w/ DARE\pub{ICML2024} \citep{43})} &20.77 & 4.41  &8.38  &11.59  &43.47  &46.28    \\

  {TA (w/ DARE)}  &21.25 & 4.81  &8.49 &11.82 &43.67 &46.42       \\
  
  {TIES (w/ DARE)} &20.98 & 4.62 &8.31 &11.47 &43.14 &46.28       \\
  
  

  \hline
  \textit{\small --Mitigating Catastrophic Forgetting}  & \\
     {\cellcolor[HTML]{FDF5F1}\textbf{\ourapproach-Clotho-AQA}}  &\cellcolor[HTML]{FDF5F1}21.87 &\cellcolor[HTML]{FDF5F1}4.92 &\cellcolor[HTML]{FDF5F1}8.46 &\cellcolor[HTML]{FDF5F1}11.52 &\cellcolor[HTML]{FDF5F1}43.55 
     &\cellcolor[HTML]{FDF5F1}46.28  \\

   {\cellcolor[HTML]{FDF5F1}\textbf{\ourapproach-Flickr30k}}  &\cellcolor[HTML]{FDF5F1}22.03 &\cellcolor[HTML]{FDF5F1}5.08 &\cellcolor[HTML]{FDF5F1}8.55 &\cellcolor[HTML]{FDF5F1}11.63 &\cellcolor[HTML]{FDF5F1}43.61 &\cellcolor[HTML]{FDF5F1}46.36     \\

   {\cellcolor[HTML]{FDF5F1}\textbf{\ourapproach-Clotho-AQA+Flickr30k}}  &\cellcolor[HTML]{FDF5F1}21.56 &\cellcolor[HTML]{FDF5F1}4.98 &\cellcolor[HTML]{FDF5F1}8.39&\cellcolor[HTML]{FDF5F1}11.38 &\cellcolor[HTML]{FDF5F1}43.34 &\cellcolor[HTML]{FDF5F1}46.02        \\
  \hline
  \end{tabular}
}
    \caption{Results for each method on two point cloud tasks. Among them, ModelNet40 is a classification task, while Objavers is a captioning task.
    }
  \label{table2_detail_P}
  \vspace{-10pt}
  \end{table*}

%% file: tables/table2_detail_AV.tex
  \begin{table*}[ht]
    
  \centering
    \renewcommand{\arraystretch}{1.15}
  \resizebox{1\linewidth}{!}{
  \begin{tabular}{l|cc|ccc|cc}

      \bottomrule

& \multicolumn{5}{c|}{\centering \textbf{3 Audio Tasks}} & \multicolumn{2}{c}{\centering \textbf{2 Video Tasks}} \\ 

\cline{2-8}

\multirow{-2}{*}{Task ($\rightarrow$)} & { \textcolor[rgb]{0.27, 0.35, 0.760} {TUT}}   
& { \textcolor[rgb]{0.27, 0.35, 0.760} {VocalSound}} 
&\multicolumn{3}{c|}{\centering\textcolor[rgb]{0.27, 0.35, 0.760} {Clotho}}  & { \textcolor[rgb]{0.27, 0.35, 0.760} {MSVD}} & { \textcolor[rgb]{0.27, 0.35, 0.760} {MSRVTT}} \\

Method ($\downarrow$)  & Acc. & Acc. & CIDEr  & SPICE  &  SPIDEr & Acc.  & Acc.\\   

\hline
      \cellcolor[HTML]{F5F5F5}{Original MLLMs}  
      &\cellcolor[HTML]{F5F5F5}22.23&\cellcolor[HTML]{F5F5F5}27.19 & \cellcolor[HTML]{F5F5F5}38.63&\cellcolor[HTML]{F5F5F5}11.98& \cellcolor[HTML]{F5F5F5}25.29&\cellcolor[HTML]{F5F5F5}48.40&\cellcolor[HTML]{F5F5F5}31.18 \\

        {\cellcolor[HTML]{FDF5F1}\textbf{\ourapproach (ours)}} 
  &\cellcolor[HTML]{FDF5F1}34.14
  &\cellcolor[HTML]{FDF5F1}42.88 &\cellcolor[HTML]{FDF5F1}38.49 &\cellcolor[HTML]{FDF5F1}11.93 &\cellcolor[HTML]{FDF5F1}25.18 &\cellcolor[HTML]{FDF5F1}48.12    &\cellcolor[HTML]{FDF5F1}30.43    \\

  \hline
  \textit{\small –Multi-Modality Retention} & & & & & & &  \\

    {NaiveMC} \pub{ACL2024} \citep{18} &29.50 &31.80 &37.56 &11.61 &24.61 &44.53 &29.31    \\
    
    {TA} \pub{ICLR23} \citep{21} &30.64 &33.12 &37.69 &11.67 &24.69 &45.61 &29.54       \\
  
  {TIES} \pub{NeurIPS23} \citep{24} &30.87 &33.42 &37.89 &11.72 &24.78 &45.88 &29.74       \\

{NaiveMC (w/ DARE\pub{ICML2024} \citep{43})}  &30.50 &32.75 &37.75 &11.66 &24.74 &45.69 &29.58    \\

  {TA (w/ DARE)}  &30.98 &33.90 &37.87 &11.69 &24.89 &45.51 &29.54       \\
  
  {TIES (w/ DARE)} &31.59 &34.45 &37.96 &11.87 &24.92 &46.07 &29.93       \\
  
  

  \hline
  \textit{\small --Mitigating Catastrophic Forgetting} & & & & & & &  \\
  {Fine-tune on Clotho-AQA} &6.98 &17.65 &30.02 &9.40 &20.04 &- &- \\
  
     {\cellcolor[HTML]{FDF5F1}\textbf{\ourapproach-Clotho-AQA}}  &\cellcolor[HTML]{FDF5F1}34.01 &\cellcolor[HTML]{FDF5F1}42.45 &\cellcolor[HTML]{FDF5F1}38.37 &\cellcolor[HTML]{FDF5F1}11.89 &\cellcolor[HTML]{FDF5F1}25.11 
     &\cellcolor[HTML]{FDF5F1}48.04 
     &\cellcolor[HTML]{FDF5F1}30.29\\

   {\cellcolor[HTML]{FDF5F1}\textbf{\ourapproach-Flickr30k}}  &\cellcolor[HTML]{FDF5F1}33.41 &\cellcolor[HTML]{FDF5F1}41.94 &\cellcolor[HTML]{FDF5F1}38.10&\cellcolor[HTML]{FDF5F1}11.81 &\cellcolor[HTML]{FDF5F1}24.98 &\cellcolor[HTML]{FDF5F1}47.74 &\cellcolor[HTML]{FDF5F1}30.05     \\

   {\cellcolor[HTML]{FDF5F1}\textbf{\ourapproach-Clotho-AQA+Flickr30k}}  &\cellcolor[HTML]{FDF5F1}33.54 &\cellcolor[HTML]{FDF5F1}41.83 &\cellcolor[HTML]{FDF5F1}37.97&\cellcolor[HTML]{FDF5F1}11.76 &\cellcolor[HTML]{FDF5F1}24.92 &\cellcolor[HTML]{FDF5F1}47.38  &\cellcolor[HTML]{FDF5F1}29.67        \\
  \hline

  \end{tabular}
}
\caption{Results for each method on three audio tasks and two video tasks. Among them, TUT, VocalSound, MSVD, and MSRVTT are the classification tasks, while Clotho is a captioning task.
    }
  \label{table2_detail_AV}
  \vspace{-10pt}
  \end{table*}

%% file: tables/table3_detail_multimodal.tex
\begin{table*}[ht]
\centering
\renewcommand{\arraystretch}{1.22}
\captionsetup{type=table}

\resizebox{1.0\linewidth}{!}{  
\begin{tabular}{l|c|ccc|ccccc}
\bottomrule
\rowcolor{mygray}
Task ($\rightarrow$) &  \textbf{ModelNet40} & \multicolumn{3}{c|}{ \textbf{MUSCI-AVQA} } & \multicolumn{5}{c}{\centering \textbf{MCUB}}  \\ 
\rowcolor{mygray}
Method ($\downarrow$) & PI-T & IA-T & VI-T &VA-T & AVI-T & AVP-T & AIP-T & VIP-T & AVIP-T  \\

\hline
\textbf{\ourapproach-Clotho-AQA} &61.98	&47.01	&51.22	&51.43 &56.08	&59.11	&65.08	&55.80	&61.08 \\

\textbf{\ourapproach-Flickr30k} &61.84	&46.92	&51.05	&51.56 &56.28	&58.90	&65.08	&55.40	&60.93 \\

\textbf{\ourapproach-Clotho-AQA+Flickr30k} &61.33	&46.48	&50.61	&51.17 &55.68	&57.93	&64.17	&55.20	&60.42
 \\

\bottomrule
\end{tabular}
}
\caption{Results of the mitigating catastrophic forgetting experiments for three \ourapproach variants on multimodal tasks with different combinations of video (V), image (I), audio (A), point cloud (P), and text (T) inputs.}
\label{table3_detail_multimodal}
\vspace{-10pt}
\end{table*}

%% file: tables/table3_detail_new.tex
  \begin{table*}[ht]

  \centering
    \renewcommand{\arraystretch}{1.2}
  \resizebox{1\linewidth}{!}{
  \begin{tabular}{l|c|ccccccc}

      \bottomrule

Task ($\rightarrow$) & \multicolumn{1}{c|}{\centering \textcolor[rgb]{0.27, 0.35, 0.760}{Clotho-AQA}} & \multicolumn{7}{c}{\centering \textcolor[rgb]{0.27, 0.35, 0.760}{Flickr30k}} \\ 

\cline{2-9}

Method ($\downarrow$)  & Acc. & CIDEr & METEOR  & BLEU-1  &  BLEU-2 & BLEU-3  & BLEU-4 &ROUGE \\   

\hline
      \cellcolor[HTML]{F5F5F5}{Original MLLMs}  
      &\cellcolor[HTML]{F5F5F5}49.40&\cellcolor[HTML]{F5F5F5}80.27	&\cellcolor[HTML]{F5F5F5}25.62	&\cellcolor[HTML]{F5F5F5}73.29	&\cellcolor[HTML]{F5F5F5}55.81	&\cellcolor[HTML]{F5F5F5}41.11	&\cellcolor[HTML]{F5F5F5}29.66	&\cellcolor[HTML]{F5F5F5}53.11
 \\
{Fine-tune on Clotho-AQA} &57.80 &- &- &- &- &- &- &-    \\
{Fine-tune on  Flickr30k} &- &94.25	&27.74	&78.27	&62.24	&47.99	&36.50	&57.04
    \\
  \hline
 {\cellcolor[HTML]{F5F5F5}\textbf{\ourapproach (ours)}}  &\cellcolor[HTML]{F5F5F5}49.28  &\cellcolor[HTML]{F5F5F5}79.56	&\cellcolor[HTML]{F5F5F5}25.56	&\cellcolor[HTML]{F5F5F5}73.11	&\cellcolor[HTML]{F5F5F5}55.62	&\cellcolor[HTML]{F5F5F5}40.83	&\cellcolor[HTML]{F5F5F5}29.49	&\cellcolor[HTML]{F5F5F5}52.82
   \\

     {\cellcolor[HTML]{FDF5F1}\textbf{\ourapproach-Clotho-AQA}}  &\cellcolor[HTML]{FDF5F1}57.71 &\cellcolor[HTML]{FDF5F1}79.72 &\cellcolor[HTML]{FDF5F1}25.51 &\cellcolor[HTML]{FDF5F1}73.04 &\cellcolor[HTML]{FDF5F1}55.48 
     &\cellcolor[HTML]{FDF5F1}40.72 
     &\cellcolor[HTML]{FDF5F1}29.33
     &\cellcolor[HTML]{FDF5F1}52.77\\

   {\cellcolor[HTML]{FDF5F1}\textbf{\ourapproach-Flickr30k}}  &\cellcolor[HTML]{FDF5F1}48.94 &\cellcolor[HTML]{FDF5F1}92.74 &\cellcolor[HTML]{FDF5F1}27.58&\cellcolor[HTML]{FDF5F1}77.93 &\cellcolor[HTML]{FDF5F1}61.69 &\cellcolor[HTML]{FDF5F1}47.36 &\cellcolor[HTML]{FDF5F1}36.02   &\cellcolor[HTML]{FDF5F1}56.27  \\

   {\cellcolor[HTML]{FDF5F1}\textbf{\ourapproach-Clotho-AQA+Flickr30k}}  &\cellcolor[HTML]{FDF5F1}57.52 &\cellcolor[HTML]{FDF5F1}92.09 &\cellcolor[HTML]{FDF5F1}27.29&\cellcolor[HTML]{FDF5F1}77.22 &\cellcolor[HTML]{FDF5F1}61.11 &\cellcolor[HTML]{FDF5F1}46.86  &\cellcolor[HTML]{FDF5F1}35.62   &\cellcolor[HTML]{FDF5F1}55.72     \\
  \hline

  \end{tabular}
}
    \caption{Results of the mitigating catastrophic forgetting experiments for each method on two \textbf{new tasks}. Among them, Clotho-AQA is a Question-Answering task, while Flickr30k is a captioning task.
    }
  \label{table3_detail_new}
  \vspace{-8pt}
  \end{table*}

%% file: tables/table_prompts.tex
\begin{table*}[ht]
\centering

\begin{tabular}{
  >{\raggedright\arraybackslash}p{1.3in}
  >{\raggedright\arraybackslash}p{0.6in}
  >{\raggedright\arraybackslash}p{3.9in}
}
\toprule

Benchmark & Modality & Prompt Template \\ 

\midrule
    &AVI-T &Based on four input entities:\textbackslash nimage <image>\textbackslash naudio <audio>\textbackslash nvideo <video>\textbackslash n \{Question\} \{Options\} Answer with the option's letter from the given choices directly.     \\
  
 &AVP-T &Based on four input entities:\textbackslash naudio <audio>\textbackslash nvideo <video>\textbackslash npoint <point>\textbackslash n \{Question\} \{Options\} Answer with the option's letter from the given choices directly. \\

MCUB &VIP-T &Based on four input entities:\textbackslash nimage <image>\textbackslash nvideo <video>\textbackslash npoint <point>\textbackslash n \{Question\} \{Options\} Answer with the option's letter from the given choices directly. \\

&AIP-T &Based on three input entities:\textbackslash nimage <image>\textbackslash naudio <audio>\textbackslash npoint <point>\textbackslash n \{Question\} \{Options\} Answer with the option's letter from the given choices directly. \\ 

&AVIP-T &Based on four input entities:\textbackslash nimage <image>\textbackslash naudio <audio>\textbackslash nvideo <video>\textbackslash npoint <point>\textbackslash n \{Question\} \{Options\} Answer with the option's letter from the given choices directly. \\ 

\midrule

&VI-T &Based on the video <video> and image <image>\textbackslash n\{Question\} \textbackslash nAnswer the question using a single word. \\

MUSIC-AVQA &VA-T &Based on the video <video> and audio <audio>\textbackslash n\{Question\} \textbackslash nAnswer the question using a single word. \\

&IA-T &Based on the image <image> and audio <audio>\textbackslash n\{Question\} \textbackslash nAnswer the question using a single word. \\

\midrule

ModelNet40 &PI-T &Based on rendered image <image> and point cloud <point>\textbackslash nWhat is this? Select from these objects: \{Options\} Answer the question using a single word.
\\
&I-T &<point>\textbackslash nWhat is this? Select from these objects: \{Options\} Answer the question using a single word. \\

\midrule
Objaverse &I-T &<point>\textbackslash nOffer a clear and concise description of this point cloud object. \\

\midrule

VocalSound \& TUT &A-T &<audio>\textbackslash nWhich of the following categories does this audio belong to? \{Options\} Answer the question using a single word.\\

Clotho &A-T &<audio>\textbackslash nDescribe this audio in detail.\\

Clotho-AQA &A-T &<audio>\textbackslash n\{Question\}\textbackslash  nAnswer the question using a single word or phrase.\\

\midrule
MSRVTT \& MSVD &V-T &<video>\textbackslash n\{Question\}\textbackslash  nAnswer the question using a single word or phrase.\\

\midrule
VQAv2 \& GQA \& POPE \& OK-VQA &I-T &<image>\textbackslash n\{Question\}\textbackslash  nAnswer the question using a single word or phrase.\\

Textvqa &I-T &<image>\textbackslash n\{Question\}\textbackslash  nReference OCR token: \{Options\}\textbackslash  nAnswer the question using a single word or phrase.\\

VizWiz &I-T &<image>\textbackslash n\{Question\}\textbackslash  nWhen the provided information is insufficient, respond with 'Unanswerable'.\textbackslash  nAnswer the question using a single word or phrase.\\

ScienceQA &I-T &<image>\textbackslash n\{Context\}\textbackslash  n\{Question\}\textbackslash  nChoose the most likely ratio. \{Options\} \\

Flickr30k &I-T &<image>\textbackslash nDescribe this image using one or more simple sentences. \\

\bottomrule

\end{tabular}
\caption{Prompt Template for different evaluation benchmarks.}
\label{table:prompts}
\end{table*}